\documentclass{article}

\usepackage[final]{corl_2022} 

\usepackage{hyperref}       
\usepackage{url}            
\usepackage{booktabs}       
\usepackage{amsfonts}       
\usepackage{nicefrac}       
\usepackage{microtype}      
\usepackage{xcolor}         
\usepackage{color}
\usepackage{amsmath}
\usepackage{graphicx}
\usepackage{amsthm}
\usepackage{amssymb}
\usepackage{algorithm}
\usepackage{algorithmic}
\usepackage{subfig}
\usepackage{chngpage}
\usepackage{wrapfig}
\usepackage{enumitem}
\usepackage{multirow}
\usepackage{adjustbox}
\usepackage{colortbl}

\title{Discriminator-Guided Model-Based Offline Imitation Learning}

\author{
Wenjia Zhang$^{1}$,
Haoran Xu$^{2}$, Haoyi Niu$^{1}$, Peng Cheng$^{3}$,\\ \textbf{Ming Li}$^{1}$, \textbf{Heming Zhang}$^{1}$, \textbf{Guyue Zhou}$^{1}$, \textbf{Xianyuan Zhan}$^{1, 4}$\thanks{Corresponding author.}\\
$^1$ Tsinghua University, Beijing, China\\
$^2$ JD Technology, Beijing, China\\
$^3$ Beijing Jiaotong University, Beijing, China\\
$^4$ Shanghai AI Laboratory, Shanghai, China\\
\texttt{zhangwj18@mails.tsinghua.edu.cn}\\
\texttt{\{ryanxhr,t6.da.thu,liming18739796090\}@gmail.com}\\
\texttt{20112041@bjtu.edu.cn}\\
\texttt{hmz@mail.tsinghua.edu.cn}\\
\texttt{\{zhouguyue,zhanxianyuan\}@air.tsinghua.edu.cn}
}

\begin{document}
\maketitle


\begin{abstract}
		Offline imitation learning (IL) is a powerful method to solve decision-making problems from expert demonstrations without reward labels. Existing offline IL methods suffer from severe performance degeneration under limited expert data.
		Including a learned dynamics model can potentially improve the state-action space coverage of expert data, however, it also faces challenging issues like model approximation/generalization errors and suboptimality of rollout data.
		In this paper, we propose the Discriminator-guided Model-based offline Imitation Learning (DMIL) framework,
		which introduces a discriminator to simultaneously distinguish the dynamics correctness and suboptimality of model rollout data against real expert demonstrations. DMIL adopts a novel cooperative-yet-adversarial learning strategy, which uses the discriminator to guide and couple the learning process of the policy and dynamics model, resulting in improved model performance and robustness.
		Our framework can also be extended to the case when demonstrations contain a large proportion of suboptimal data. Experimental results show that DMIL and its extension achieve superior performance and robustness compared to state-of-the-art offline IL methods under small datasets.
	\end{abstract}

	
	
		\section{Introduction}
	
	Offline imitation learning (IL) that trains a policy from expert demonstrations without additional online environment interactions has become an attractive solution for many real-world decision-making applications, such as robotic manipulation~\cite{robotic} and autonomous driving~\cite{bc,IL-AD}, etc. It bypasses several major obstacles in practice, such as the difficult reward function design~\cite{ng1999policy} as in reinforcement learning (RL) approaches, and the requirement of simulation or real-world system interactions during model training as in online IL methods~\cite{dagger,gail,DAC,maximum,valuedice}, which can be costly or dangerous.
	
	Despite these desirable features, the performance of offline IL methods heavily depends on the size and quality of demonstration data. Due to its supervised learning nature, learning an IL policy in parts of the state space not covered by expert data could make arbitrary mistakes, which leads to severe compounding errors. This phenomenon, called \textit{covariate shift}~\cite{limit,feedback,covariate}, is a core issue in IL and greatly hurts the policy generalization capability. In practice, collecting a large number of expert demonstrations can be costly or infeasible. The reduction in data size coupled with the narrow expert data distribution can lead to limited state space coverage, causing poor policy performance. On the other hand, involving non-expert suboptimal offline demonstration data although can potentially improve state-action space coverage, is shown in previous studies~\cite{oril,mandlekar2022matters} to result in reduced performance in traditional offline IL methods like behavior cloning (BC)~\cite{bc}. Many of these problems can be alleviated in the online IL setting, either by interactively querying an expert to collect more data~\cite{dagger,safedagger,hgdagger}, or by resorting to inverse reinforcement learning (IRL) to learn a rewards function or match the state-action distribution induced by the expert policy~\cite{gail,maximum,valuedice,variational}. However, such treatments do not apply to the offline setting, since additional environment interaction is not possible. Moreover, utilizing additional suboptimal offline data through offline IRL approaches~\cite{oril,konyushkova2020semi} also shows inferior performance compared with online IRL counterpart methods, due to the involvement of offline RL sub-problems that is prone to training instability and bootstrapping error accumulation~\cite{kumar2019stabilizing,dwbc}.
	Hence, the ability to leverage limited expert data for robust policy learning remains to be a key challenge for the successful real-world deployment of offline IL methods.
	
\begin{figure}[t]
		\small 
		\begin{minipage}[b]{.33\linewidth}
			\centerline{\includegraphics[width=4.1cm]{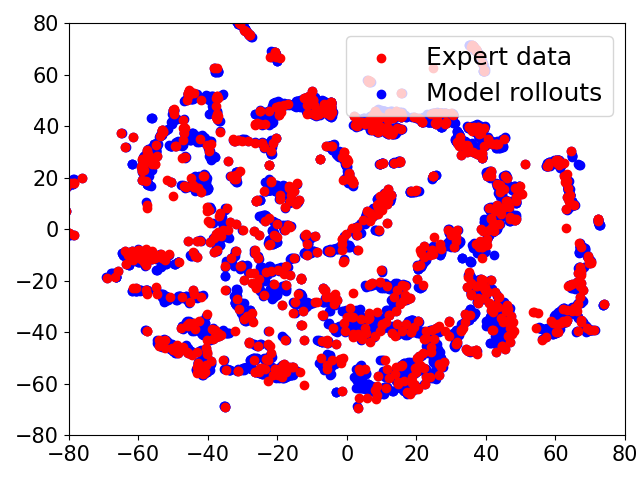}}
			\centerline{(a) Good policy and model}
			\centerline{\includegraphics[width=4.1cm]{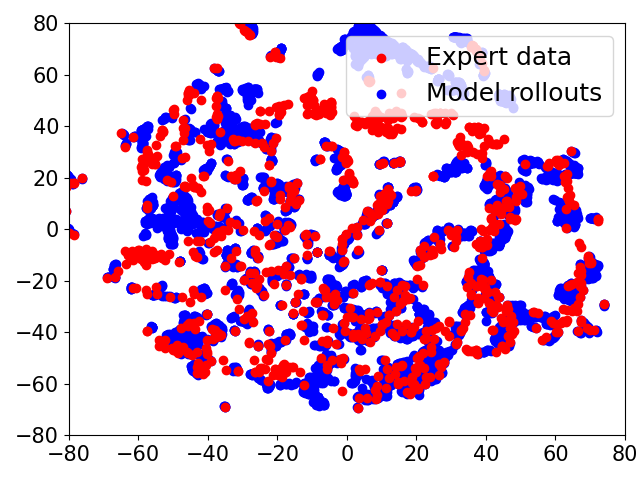}}
			\centerline{(c) Bad policy and good model }
		\end{minipage} 
		\hfill
		\begin{minipage}[b]{.33\linewidth}
			\centerline{\includegraphics[width=4.1cm]{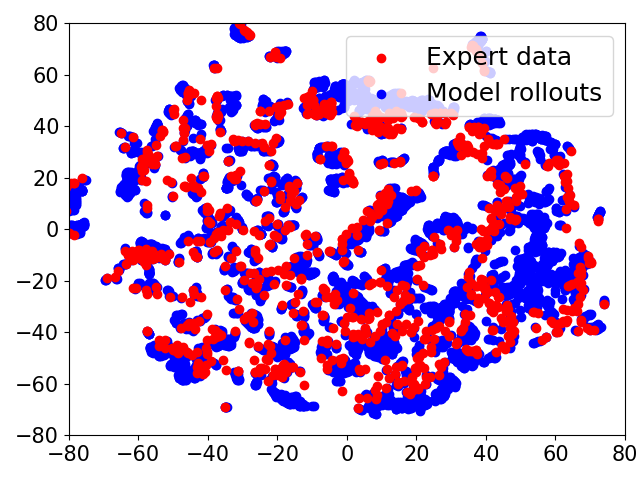}}
			\centerline{(b) Good policy and bad model}
			\centerline{\includegraphics[width=4.1cm]{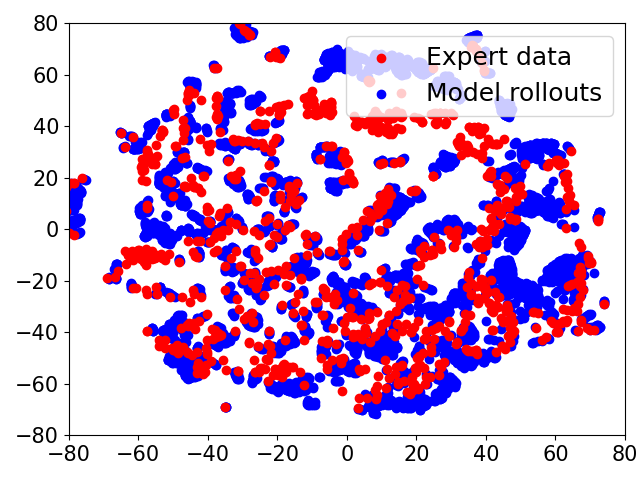}}
			\centerline{(d) Bad policy and model }
		\end{minipage} 
		\hfill
		\begin{minipage}[b]{.29\linewidth}
			\centerline{\includegraphics[width=3.4cm]{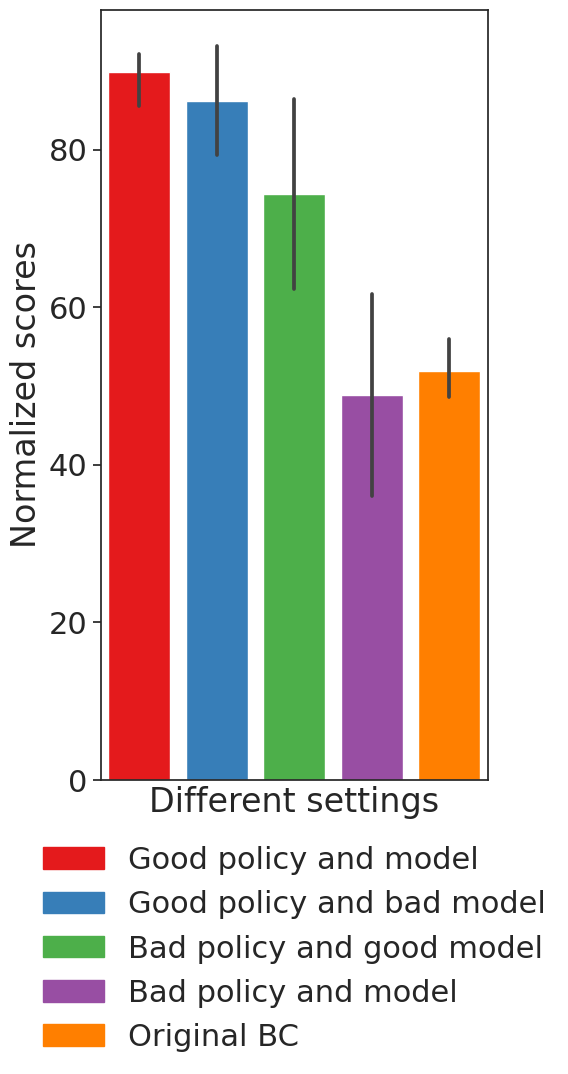}}
			\centerline{(e) Evaluation scores}
		\end{minipage} 
		\caption{\small Empirical observations on the impacts of involving dynamics model rollouts in BC. (a)-(d) TSNE visualizations of expert data and dynamics model rollouts under different BC policies on MuJoCO Hopper task with only 20,000 expert data transitions (2\% of the D4RL~\cite{d4rl} Hopper-expert dataset). The good policy and dynamics model are two-layer MLPs with 256 hidden units, and are trained until convergence. The bad policy and model are trained with fewer steps, and the hidden layers of the latter are reduced to 128 units.
		It can be observed that model rollouts under well-learned policy and dynamics model align well with the expert data, while noticeable discrepancies are observed when the policy or the model is problematic. (e) shows the final performance of BC policy trained with 1:1 expert and model rollout data under the four cases in (a)-(d). It is found that under small expert datasets, including a dynamics model in many cases is beneficial, but the quality of rollout policy and dynamics model could have great impact on the final policy performance.
		}
		\label{tsne}
	\end{figure}

The sample efficiency requirement for offline IL methods reminds us of the success of model-based approaches in the online and offline RL domains~\cite{MBPO,MOPO,modelRL,zhan2021deepthermal,zhan2021model}. Dynamics models learned from the data can greatly supplement the limited expert data to improve state-action space coverage, leading to potentially improved policy performance and generalizability~\cite{MOPO,zhan2021model,milo,ode}. However, adopting a model-based approach in offline IL is still an underexplored area~\cite{milo,ode,nips2020}. Many existing methods bear some limitations, such as requiring an additional suboptimal dataset~\cite{milo} or a low-fidelity simulator~\cite{nips2020} for training, or fully trusting the learned dynamics model~\cite{ode}.
	The key challenges of introducing a learned dynamics model in IL policy learning is twofold (see Figure~\ref{tsne} for an empirical illustration): 1) the learned dynamics model has approximation/generalization errors, directly using model rollouts for imitation learning can be problematic; 2) using the learned policy as the rollout policy may generate suboptimal data, causing performance degeneration that similar to the case of learning with suboptimal data in IL~\cite{oril,dwbc}.
	In model-based RL, the second problem is less severe, as the reward function can be used to distinguish the optimality of data. However, this is typically not possible in IL settings.	
	
	In this work, we develop a novel model-based offline IL framework to tackle the above challenges. We introduce a discriminator to simultaneously distinguish the dynamics discrepancy and suboptimality of the model rollout data against the real expert demonstrations.
	This gives rise to a special cooperative-yet-adversarial ``three-party game''. 
	Both the dynamics model and the policy provide information as inputs to the discriminator, while also challenging it to establish worst-case error minimization. Under this design, the discriminator can use more information to make better judgment on the dynamics correctness and optimality of the rollout data, and the worst-case optimization scheme also substantially improves the robustness of all three models (policy, dynamics model and discriminator).		
	Interestingly, 
	we can show that this design leads to new IL policy and dynamics model learning objectives, where the outputs of the discriminator sever as weights in their original loss functions. Moreover, the resulting algorithm can be efficiently solved in a simple supervised learning manner, which avoids explicitly solving the complex min-max optimization problems as in adversarial learning~\cite{goodfellow2014generative,carlini2019evaluating}. We thus term our algorithm Discriminator-guided Model-based Imitation Learning (DMIL). 
	Our proposed framework can also be extended to the offline IL setting that involves limited expert and a larger proportion of unknown quality, potentially suboptimal data~\cite{oril,dwbc}. This can be achieved  by simply introducing the second discriminator to contrast the expert and suboptimal data, which we refer this variant as Dual-Discriminator guided Model-based Imitation Learning (D2MIL).
	Through extensive experiments on D4RL benchmarks~\cite{d4rl} and real-world robotic tasks, we show that both DMIL and D2MIL achieve superior performance and robustness against state-of-the-art methods under small datasets. These promising results demonstrate the potential of adopting model-based learning in real-world  offline IL applications under limited expert demonstrations.

	\section{Method} 
	\begin{figure}[t]
		\centering
		\begin{adjustwidth}{-0.15cm}{-0.15cm}
			\subfloat[\small DMIL: learning from expert demonstrations only]{\includegraphics[height=4cm]{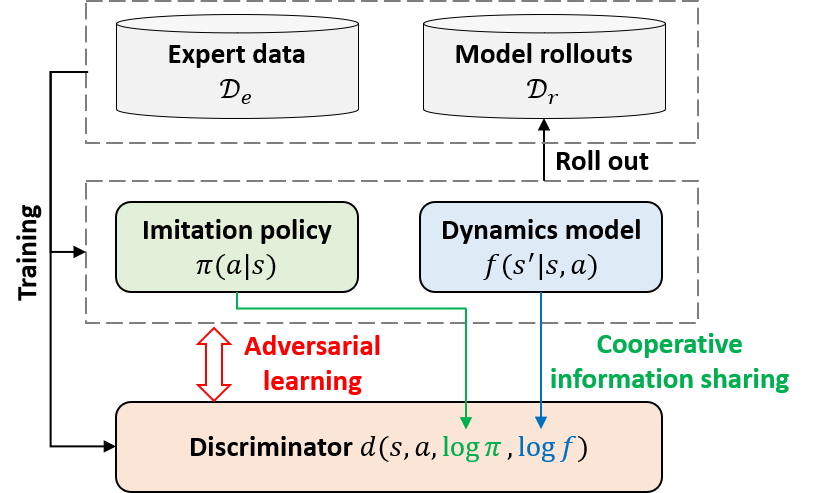}}\quad
			\subfloat[\small D2MIL: learning from both expert and suboptimal data]{\includegraphics[height=4cm]{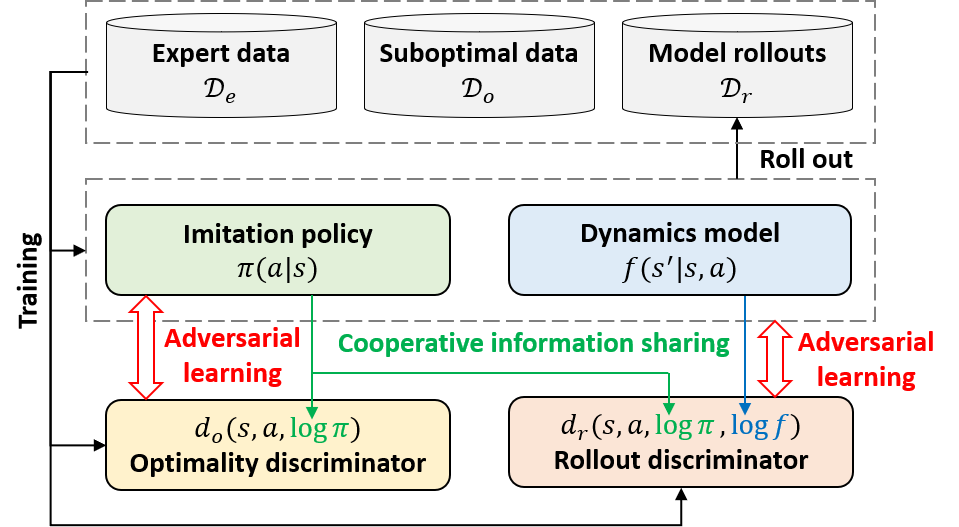}}
		\end{adjustwidth}
		\caption{\small Illustration of the proposed model-based offline IL framework DMIL and its extension D2MIL}
		\label{fig:framework}
	\end{figure}

	\subsection{Problem Setting}
	We consider the fully observed Markov Decision Process (MDP) setting, which 
	can be described as $\mathcal{M}=(\mathcal{S},\mathcal{A},P,d_0,r,\gamma)$, where $\mathcal{S}$ and $\mathcal{A}$ are the state and action space, respectively, $P(s'|s,a)$ is the transition probability, $d_0(s)$ is the initial state distribution,
	$r(s,a)$ is the reward function, and $\gamma\in[0,1]$ is the discount factor. Under offline IL setting, we have 
	an expert dataset $\mathcal{D}_e=\{(s_i,a_i,s_i')\}_{i=1}^N$ collected from some expert policy $\pi_e$. Our goal is to learn a policy $\pi(a|s)$ to minimize its gap with the expert policy $\pi_e$. In the simplest case, behavior cloning (BC) trains the policy by minimizing the negative log-likelihood of the observed expert actions:
	\begin{equation}
		\min\limits_{\pi} \mathcal{L}_{\pi} := \mathbb{E}_{(s,a)\sim \mathcal{D}_e}[-\log\pi(a|s)]\label{eq:bc}
	\end{equation}
	
	\subsection{Discriminator-Guided Model-Based Imitation Learning (DMIL)}\label{sec:method_dmil}
	Traditional offline IL methods like BC suffer greatly from covariate shift under small expert datasets due to extremely sparse state space coverage of data. Our idea is to mitigate this issue by involving dynamics model rollouts while also carefully handling these potentially problematic data through the guidance of an additional discriminator in a coupled and cooperative-yet-adversarial learning process. Figure~\ref{fig:framework} provides an illustration of the proposed  DMIL framework as well as its extension D2MIL. 
	
	\textbf{Incorporating the Dynamics Model. } Model-based approaches have been widely adopted in RL to improve sample efficiency and shows good performance and generalization ability in recent offline RL studies~\cite{MOPO,zhan2021deepthermal,zhan2021model}. In our work, we introduce a probabilistic dynamics model implemented using a neural network that outputs a Gaussian distribution over the difference between the current and next state, i.e., $f(s'|s,a)=\mathcal{N}(s+\mu_{\theta_f}(s,a),\Sigma_{\theta_f}(s,a))$, where $\mu_{\theta_f}(s,a)$ and $\Sigma_{\theta_f}(s,a)$ are the parameterized mean and diagonal covariance matrix.
	We predict the difference of states rather than the next states as it has been shown in past studies~\cite{MBPO,MOPO} to yield better dynamics predictions.
	The dynamics model can be learned using the following maximum log-likelihood objective:
	\begin{equation}
		\min\limits_{f} \mathcal{L}_{f} := \mathbb{E}_{(s,a,s')\sim \mathcal{D}_e}[-\log f(s'|s,a)]\label{eq:f}
	\end{equation}

	\textbf{Cooperative-yet-Adversarial Learning Scheme. } Directly using the rollout data $\mathcal{D}_r$ generated by the learned BC policy $\pi$ and dynamics model $f$ in subsequent imitation learning can be problematic. Under small datasets, it is usually difficult to obtain an accurate dynamics model, and the rollouts from a less-well learned policy can be suboptimal compare with the true expert data. 
	To solve this issue, we use a discriminator $d$ to measure dynamics discrepancy and suboptimality in rollout data $\mathcal{D}_r$. Moreover, we introduce a special cooperative-yet-adversarial learning scheme, and use the discriminator as a bridge to couple the learning process of $\pi$, $f$ and $d$.
	The key idea is to first include the element-wise loss information from both policy $\pi$ and dynamics model $f$ (i.e., $\log\pi$ and $\log f$) into the input of the discriminator (i.e., $d(s,a,\log\pi(a|s),\log f(s'|s,a))$) to establish cooperative information sharing. 
	And then make $\pi$ and $f$ challenge $d$ to establish adversarial learning. 
	This leads to a special learning objective for the discriminator $d$, which can be expressed as:
	\begin{equation}
		\begin{aligned}
			\underset{d}{\min}\, \underset{\pi,f}{\max}\; \mathcal{L}_d := &\underset{(s,a,s')\sim\mathcal{D}_{e}}{\mathbb{E}}[-\log d(s,a,\log\pi(a|s),\log f(s'|s,a))]+\\
			&\underset{(s,a,s')\sim\mathcal{D}_{r}}{\mathbb{E}}[-\log(1-d(s,a,\log\pi(a|s),\log f(s'|s,a)))] \label{eq:d}
		\end{aligned}
	\end{equation}
	This design has a number of attractive properties. First, element-wise loss information from $f$ and $\pi$ reflects the confidence of these models on the rollout data. Suppose $f$ and $\pi$ are well-learned, then they will assign high probabilities (large $\log \pi$ and $\log f$) on good rollouts with reasonable dynamics and expert-like samples. This can provide valuable information to facilitate the judgment of the discriminator. Second, the adversarial component forms a GAN-like problem~\cite{goodfellow2014generative}, where $\pi$ and $f$ jointly serve as a generator to challenge the discriminator. This will force the discriminator to minimize the worst-case error~\cite{carlini2019evaluating,goodfellow2014explaining}, which makes its robustness significantly improved. In return, a stronger $d$ can better guide the learning of $\pi$ and $f$ to further improve their performance and make better use of the generalization power of the dynamics model.
	Consequently, this cooperative-yet-adversarial learning scheme enables coupling among policy, dynamics model and discriminator, which can potentially lead to boosted performance for all three models.

	\textbf{Loss Correction for Policy and Dynamics Model. } Jointly solving Eq.(\ref{eq:d}) together with minimization problems in Eq.(\ref{eq:bc}) and (\ref{eq:f}) can be rather complex. As both $\pi$ and $f$ appear in the input of the discriminator, $d$ becomes a functional of $\pi$ and $f$ (i.e., function of a function). Eq.(\ref{eq:d}) is a functional min-max optimization problem, which is itself quite challenging to solve. Fortunately, based on calculus of variation~\cite{gelfand2000calculus} and the analysis method introduced in DWBC \cite{dwbc}, we can avoid directly solving this complex functional min-max optimization problem by introducing discriminator-dependent loss correction terms $\mathcal{L}_{\pi}^{corr}$ and $\mathcal{L}_{f}^{corr}$ on the losses of policy $\mathcal{L}_{\pi}$ and dynamics model $\mathcal{L}_{f}$. In this way, $\pi$, $f$ and $d$ can be efficiently learned by solving three simple minimization problems: $\min_{\pi} \alpha_{\pi}\cdot\mathcal{L}_{\pi}$ $+\mathcal{L}_{\pi}^{corr}$, $\min_{f} \alpha_f\cdot\mathcal{L}_{f}+\mathcal{L}_{f}^{corr}$ and $\min_{d} \mathcal{L}_d$, where $\alpha_{\pi}$, $\alpha_f \geq 1$ are weight factors for the original losses of $\pi$ and $f$.
	In the follows, we briefly describe the essential steps of deriving $\mathcal{L}_{\pi}^{corr}$ and $\mathcal{L}_{f}^{corr}$, and provide detailed derivations in Appendix~\ref{app:derivation}. 
	The outline of DMIL is presented in Appendix B.1.
	
	Denote $x=(s,a,s')$ and $\Omega_{sas'}$ as its domain. Note that the functional $\mathcal{L}_d(d,\log\pi,\log f)$ can be written as the integral of a new functional $F(x,\log\pi, \log f)$ with the following form:
	\begin{equation}
		\mathcal{L}_d=\int_{\Omega_{sas'}}[P_{D_e}(x)\cdot(-\log d)+P_{D_r}(x)\cdot(-(1-\log d))]\mathrm{d}x \triangleq \int_{\Omega_{sas'}} F(x,d,\log\pi, \log f) \mathrm{d}x
		\label{eq:functional}
	\end{equation}
	where 
	we slightly abuse the notations and write the output of $d(s,a,\log\pi(a|s),\log f(s'|s,a))$ as $d$ and $F(x,d,\log\pi, \log f)$ as $F$ hereafter; $P_{D_e}$ and $P_{D_r}$ are distributions of $x$ in $\mathcal{D}_e$ and $\mathcal{D}_r$. To simplify the analysis, we focus on the inner maximization problem in Eq.(\ref{eq:d}). According to calculus of variation, maximizing $\mathcal{L}_d$ with respect to function $\pi$ and $f$ requires to find the extrema of $\mathcal{L}_d$, which can be achieved by solving the following associate Euler-Lagrangian equations:
	\begin{equation}
		\left\{\begin{array}{l}
			F_{\pi}-\frac{\partial}{\partial x} F_{\frac{\partial \pi}{\partial x}}=F_{\pi}=0 \\
			F_{f}-\frac{\partial}{\partial x} F_{\frac{\partial f}{\partial x}}=F_{f}=0
		\end{array}\right.\label{eq:euler}
	\end{equation}
	
	where $F_y$ stands for $\frac{\partial F}{\partial y}$. Let $\theta_{\pi}$ and $\theta_f$ denote the network parameters of policy $\pi$ and dynamics model $f$. Using the analysis on policy $\pi$ as an example. Assuming $F$ and $d$ are continuously differentiable with respect to $d$ and $\log\pi$ respectively, from the first equation in Eq.(\ref{eq:euler}), we have $F_{\pi}\cdot \frac{\partial \pi}{\partial \theta_{\pi}}=\frac{\partial F}{\partial d} \cdot \frac{\partial d}{\partial \log\pi} \cdot \frac{\partial \log\pi}{\pi}\cdot \frac{\partial \pi}{\partial \theta_{\pi}}=\frac{\partial d}{\partial \log\pi} \cdot\frac{\partial F}{\partial d}\cdot\nabla_{\theta_{\pi}}\log\pi =0$. As $d$ is determined by the outer minimization problem of Eq.(\ref{eq:d}), thus $\frac{\partial d}{\partial \log\pi}$ is not obtainable by solely inspecting the inner maximization problem. To ensure the previous equation hold, we can instead consider a relaxed condition by letting $\frac{\partial F}{\partial d}\cdot\nabla_{\theta_{\pi}}\log\pi =0$. The integration of this new condition is still 0
	($\int_{\Omega_{sas'}}\frac{\partial F}{\partial d}\cdot\nabla_{\theta_{\pi}}\log\pi \mathrm{d}x=0$), 
	which leads to the following tractable condition:
	\begin{equation}
		-\underset{(s,a,s')\sim\mathcal{D}_e}{\mathbb{E}}\left[-\frac{1}{d}\cdot\nabla_{\theta_{\pi}}\log\pi\right]+ \underset{(s,a,s')\sim\mathcal{D}_r}{\mathbb{E}}\left[-\frac{1}{1-d}\cdot\nabla_{\theta_{\pi}}\log\pi\right] = 0
	\end{equation}
	Above can be equivalently perceived as the first-order optimality condition of minimizing the following corrective loss term $\mathcal{L}_{\pi}^{corr}$ for policy $\pi$:
	\begin{equation}
		\mathcal{L}_{\pi}^{corr}=\underset{(s,a,s')\sim\mathcal{D}_e}{\mathbb{E}} \left[\frac{1}{d}\cdot\log\pi(a|s)\right]- \underset{(s,a,s')\sim\mathcal{D}_r}{\mathbb{E}} \left[\frac{1}{1-d}\cdot\log\pi(a|s)\right]\label{eq:corr_pi}
	\end{equation}
	
	Similarly, we can obtain the corrective loss term $\mathcal{L}_{f}^{corr}$ for dynamics model $f$ as:
	\begin{equation}
		\mathcal{L}_{f}^{corr}=\underset{(s,a,s')\sim\mathcal{D}_e}{\mathbb{E}} \left[\frac{1}{d}\cdot\log f(s'|s,a)\right]- \underset{(s,a,s')\sim\mathcal{D}_r}{\mathbb{E}} \left[\frac{1}{1-d}\cdot\log f(s'|s,a)\right]\label{eq:corr_f}
	\end{equation}

		\subsection{Extensions to Scenarios with Additional Suboptimal Dataset}
	The DMIL framework can be easily extended to IL scenarios with a small expert dataset $\mathcal{D}_e$ and a larger dataset $\mathcal{D}_o$ sampled from one or multiple potentially suboptimal policies~\cite{oril,dwbc,demodice}. Under this setting, we can add a second optimality discriminator $d_o$ in additional to the original rollout discriminator in DMIL (referred as $d_r$ in this setting), 	
	dedicated to differentiate between expert and suboptimal samples in both $\mathcal{D}_o$ and $\mathcal{D}_r$. We follow \citet{dwbc} to adopt a positive-unlabeled (PU) learning~\citep{pu-learn} objective for $d_o$, and also introduce a second pair of adversarial relationship between $\pi$ and $d_o$. 
	PU-learning enables learning from positive (expert data $\mathcal{D}_e$) and unlabeled data ($\mathcal{D}_o\cup \mathcal{D}_r$ in our case) with a hyperparameter $\eta$ to capture the proportion of positive samples to unlabeled samples.
	\begin{equation}
		\begin{aligned}
			&\underset{d_o}{\min}\,\underset{\pi}{\max} \ \mathcal{L}_{d_{o}} := \eta \underset{(s, a) \sim \mathcal{D}_{e}}{\mathbb{E}}[-\log d_o(s, a, \log\pi(a|s))] +\\
			&\underset{(s, a) \sim \mathcal{D}_{o}\cup \mathcal{D}_r}{\mathbb{E}}[-\log (1-d_o(s, a,\log\pi(a|s)))]
			-\eta \underset{(s, a) \sim \mathcal{D}_{e}}{\mathbb{E}}[-\log (1-d_o(s, a, \log\pi(a|s)))] \label{eq:d_d2mil}
		\end{aligned}
	\end{equation}
	Similar to the derivation in previous section, when jointly solving above functional min-max optimization problem together with Eq.(\ref{eq:bc})-(\ref{eq:d}), we can obtain the following updated corrective loss term for policy $\pi$, which now depends on outputs of both discriminators $d_o$ and $d_r$, with $\beta_o$ and $\beta_r$ being the weight parameters for the two discriminators. 
	We term this extension as Dual-Discriminator guided Model-based Imitation Learning (D2MIL). Complete derivation can be found in Appendix~\ref{app:derivation}. 
	\begin{align}
		\mathcal{L}_{\pi}^{corr}=&\underset{(s, a,s') \sim \mathcal{D}_e}{\mathbb{E}}\left[\left(\frac{\beta_{o} \eta}{d_{o}\left(1-d_{o}\right)}+\frac{\beta_{r}}{d_{r}}\right)\cdot \log\pi(a|s) \right]-\underset{(s, a, s') \sim \mathcal{D}_o}{\mathbb{E}}\left[\left(\frac{\beta_{o} }{1-d_{o}}-\frac{\beta_{r}}{d_{r}}\right)\cdot \log\pi(a|s) \right] \notag\\
		&-\underset{(s, a, s') \sim \mathcal{D}_r}{\mathbb{E}}\left[\left(\frac{\beta_{o}}{1-d_{o}}+\frac{\beta_{r} }{1-d_{r}}\right)\cdot \log\pi(a|s) \right] \label{eq:corr_d2mil}
	\end{align}


\section{Experiments}\label{sec:exp}
	We evaluate our methods against offline IL baseline methods on both D4RL benchmark datasets~\cite{d4rl} and a real-world wheel-legged robot. Our methods achieve superior performance and robustness compared with baselines, especially under small datasets. Experiment setups and results are described below.
	Ablation study on the impact of different design elements of DMIL can be found in Appendix \ref{app:add_exps_ablation}. Implementation details and extra comparative results are reported in Appendix~\ref{app:details} and \ref{app:add_exps}.

\subsection{Experiment Setup}
\textbf{Baselines.}
	We compare DMIL with 5 baselines: 1) BC: vanilla BC~\cite{bc}; 2) BC+d: learns a dynamics model alongside BC to generate rollouts, and the policy is trained on both expert and rollout data; 3) 2-phase BC+d: first pretrains the dynamics model and a BC policy on expert data, then uses BC+d to fine-tune the policy;
	4) DWBC+d: we use a pretrained dynamics model and a BC policy to generate the suboptimal dataset required in DWBC, and then run DWBC to learn the policy;
	5) ValueDICE: we implement an offline version of the original ValueDICE~\cite{valuedice}, which uses a learned dynamics model to serve as the online sampling environment;
	6) IQ-Learn~\cite{iqlearn}: a recent IL method that learns Q function to implicitly represent the policy, and can work offline. 
	For D2MIL, we compare it with BC trained on expert data only (BC-exp) and on all data (BC-all), as well as two recent methods ORIL~\cite{oril}, DemoDICE~\cite{demodice} and DWBC~\cite{dwbc} which are designed for the same problem setting.

\textbf{Simulation Tasks.}
	We conduct the experiments on the widely-used D4RL~\cite{d4rl} MuJoCo expert/medium datasets and the more complex Adroit human datasets (Pen, Hammer, Door).
	To investigate the impact of sample size on model performance, we randomly sample certain proportions of transitions from MuJoCo expert datasets to construct a set of much smaller datasets for evaluation. 

\begin{wrapfigure}[9]{r}{4.1cm} 
\includegraphics[width=4cm]{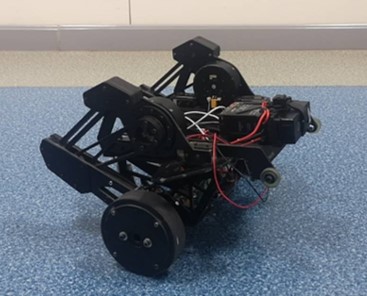}\
\caption{Wheel-legged robot}\label{robot}
\end{wrapfigure}
\textbf{Real-world Robotic Tasks.}
	We also experiment on a real-world robot which stands on a pair of wheels to get balanced, as shown in Figure \ref{robot}. The states of robot are composed of its forward tilt angle $\theta$, displacement $x$, angular velocity $\dot{\theta}$ and linear velocity $\dot{x}$. The robot is controlled by the torque $\tau$ of motors at two wheels. We evaluate our method on two tasks: (1) \textbf{Standing still}: keep the robot balanced and not fall down; (2) \textbf{Moving straight}: keep the robot balanced and move forward with a target velocity $v$. The dataset for these tasks are collected from very few human demonstrations (10,000 transitions from about 50s human control at a sampling frequency of 200Hz).


\begin{table}[t]
		\caption{\small Normalized scores for models trained on different proportion of D4RL MuJoCo-expert datasets and Adroit-human tasks. Results are averaged over 3 random seeds.}
		\label{table1}
		\small
		\begin{adjustbox}{center}
			\begin{tabular}{cccccccccc}
				\toprule
				& Ratio & BC              & BC+d &2-phase BC+d      & \textcolor{black}{DWBC+d} & ValueDICE &IQ-Learn  & DMIL            \\ \midrule
				\multirow{3}{*}{Hopper}& 100\%    & 95.06$\pm$20.38  & 106.78$\pm$4.4  & \textbf{110.59$\pm$0.63}  & \textcolor{black}{96.96$\pm$18.15} & 60.34$\pm$10.12    & 25.49$\pm$5.34                              & \textbf{110.22$\pm$1.22} \\
				& 10\%    & 83.52$\pm$30.58 & 100.59$\pm$13.21 &104.35$\pm$9.44 &  \textcolor{black}{91.52$\pm$24.81}  & 58.77$\pm$10.45    &25.16$\pm$6.69                  &\textbf{111.56$\pm$1.51} \\
				& 5\%    & 73.35$\pm$37.04 & 94.82$\pm$19.72 &99.66$\pm$14.98  &  \textcolor{black}{88.35$\pm$28.16}  & 44.94$\pm$13.71    &4.58$\pm$0.51                              & \textbf{111.14$\pm$1.83} \\
				& 2\%    & 53.54$\pm$36.89 & 61.57$\pm$30.18 &88.24$\pm$25.63 &  \textcolor{black}{81.70$\pm$32.27}    & 31.38$\pm$12.84    &3.72$\pm$0.56                        & \textbf{108.51$\pm$3.88} \\ \midrule
				\multirow{3}{*}{Halfcheetah} & 100\%    &91.95$\pm$1.24  &89.23$\pm$1.35  & 91.48$\pm$0.33   & \textcolor{black}{83.75$\pm$6.57}   & 56.07$\pm$5.33     &38.12$\pm$9.96                            &\textbf{93.34$\pm$1.29}  \\
				& 10\%    & 90.64$\pm$2.21  & 89.71$\pm$2.88 &71.27$\pm$19.33   &  \textcolor{black}{77.48$\pm$12.97}  & 48.77$\pm$8.30     &18.36$\pm$16.09                           & \textbf{92.69$\pm$1.82}  \\
				& 5\%    & 82.90$\pm$11.71 & 76.40$\pm$16.94  &70.89$\pm$23.06 &  \textcolor{black}{65.76$\pm$20.55}   & 30.61$\pm$6.98    &7.12$\pm$6.77                            & \textbf{90.18$\pm$4.43}  \\
				& 2\%    & 23.58$\pm$16.36 & 21.48$\pm$16.86  &57.48$\pm$25.63 &   \textcolor{black}{30.10$\pm$22.27}  & 17.47$\pm$7.63    &1.63$\pm$1.37                            & \textbf{76.87$\pm$15.31} \\ \midrule
				\multirow{3}{*}{Walker2d}  & 100\%    &107.35$\pm$2.29  & 106.82$\pm$1.33  &\textbf{108.15$\pm$0.27}  &  \textcolor{black}{103.92$\pm$6.53} & 86.42$\pm$11.20    & 100.96$\pm$1.23                             & 107.65$\pm$0.37 \\
				& 10\%    & 105.36$\pm$4.38 & \textbf{107.61$\pm$1.14}  &106.40$\pm$1.96 &  \textcolor{black}{91.17$\pm$25.05}   & 86.76$\pm$13.04    &73.65$\pm$12.64                              & \textbf{107.62$\pm$0.83} \\
				& 5\%    & 103.21$\pm$7.81 & 105.42$\pm$3.93 &104.51$\pm$4.54  &  \textcolor{black}{89.78$\pm$24.81}   & 83.51$\pm$12.96    &59.47$\pm$23.17                             & \textbf{107.89$\pm$0.71} \\
				& 2\%    & 58.34$\pm$35.86 & 60.64$\pm$35.10 &86.71$\pm$21.20 &  \textcolor{black}{65.19$\pm$36.27}   & 78.84$\pm$23.16    &34.19$\pm$20.11                              & \textbf{105.55$\pm$4.42} \\ 
				\midrule
				\multicolumn{2}{c}{pen-human}      & 57.91$\pm$55.05  & 7.27$\pm$15.87 &\textbf{68.57$\pm$53.57} &   \textcolor{black}{18.61$\pm$26.46}     & 52.51$\pm$19.58    & 4.94$\pm$11.51                        & \textbf{67.56$\pm$57.87}  \\ \midrule
				\multicolumn{2}{c}{hammer-human}    & 1.05$\pm$1.01  & 1.18$\pm$1.25 &1.64$\pm$1.30 &  \textcolor{black}{0.67$\pm$0.64}   & 1.12$\pm$0.64    & 0.37$\pm$0.13                           & \textbf{2.06$\pm$1.91} \\ \midrule
				\multicolumn{2}{c}{door-human}    & 0.47$\pm$0.65 & 0.16$\pm$0.29 &0.94$\pm$1.24 &  \textcolor{black}{0.01$\pm$0.21} & 0.22$\pm$0.01    & -0.28$\pm$0.01                          & \textbf{6.06$\pm$7.56} \\
				\bottomrule			
			\end{tabular}
		\end{adjustbox}
	\end{table}
	
\subsection{Results}
\textbf{Comparative Evaluation on D4RL Benchmarks. }
	The comparative results are presented in Table \ref{table1}. 
	We can see that in many tasks, na\"ively incorporating dynamics model with BC only leads to marginal improvement.
	This is due to the lack of discrimination on the quality of rollout data. 2-phase BC+d that use a pretrained, high quality dynamics model and rollout policy in some cases can result in improved performance under small dataset.
	Besides, offline ValueDICE performs poorly owing to its reliance on accurate online interaction. IQ-Learn performs badly on the continuous control tasks with high-dimensional state-action space.
	For DWBC+d, we can see that simply incorporating rollouts from dynamics model as suboptimal dataset in DWBC brings no benefit to dynamics model learning, and insufficient leverage of information in the limited data, which leads to substaintial performance drop with smaller dataset.
	By contrast, our method achieves the best performance in almost all tasks with small variance. Most importantly, we find DMIL performs surprisingly well under small datasets while other baselines suffer from severe performance degeneration. It achieves comparable performance even if the training data is reduced to 5\% or 2\% of its original size.

\begin{figure}[t] 
\centering
\begin{adjustwidth}{-0cm}{-0cm}
	\subfloat[\small Standing still]{\includegraphics[height=3.1cm]{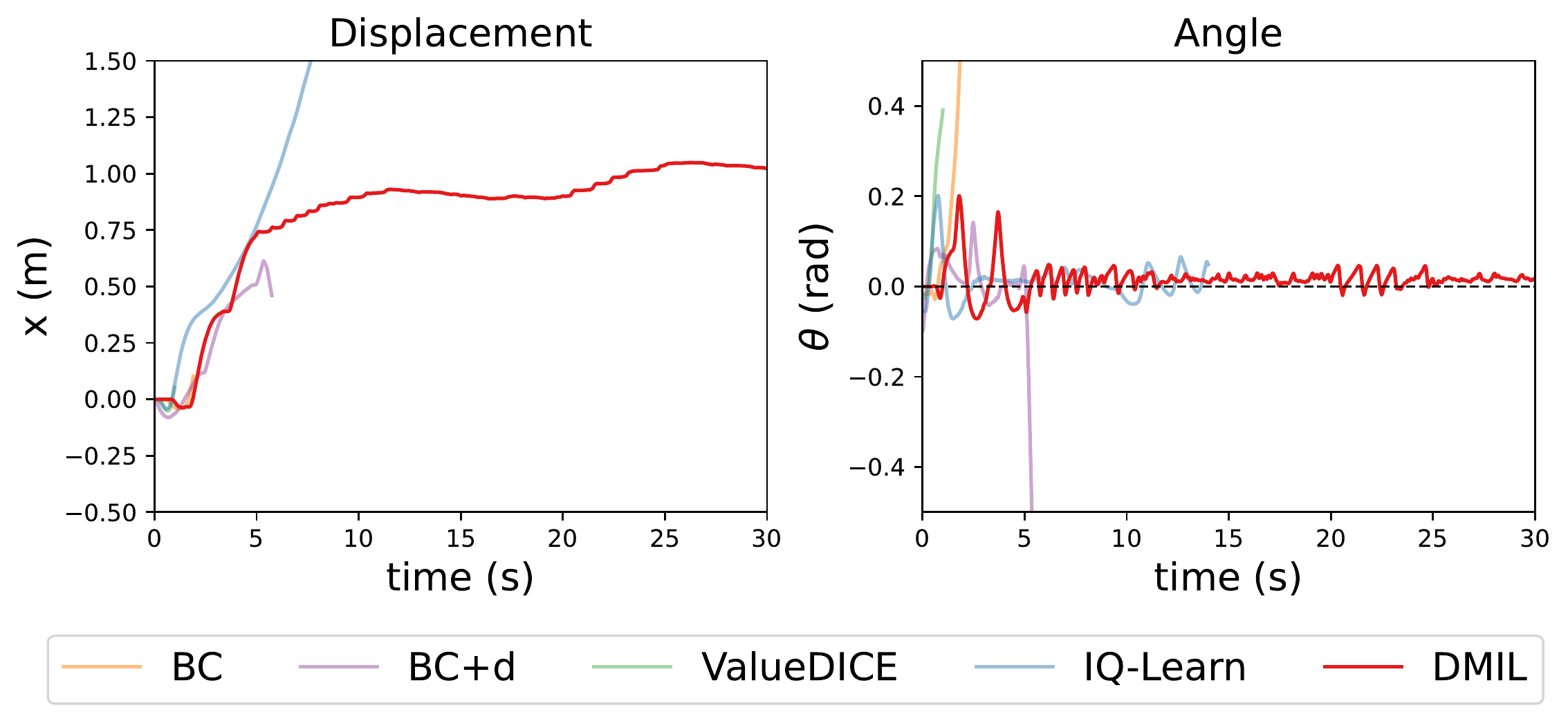}}\;
	\subfloat[\small Moving straight]{\includegraphics[height=3.1cm]{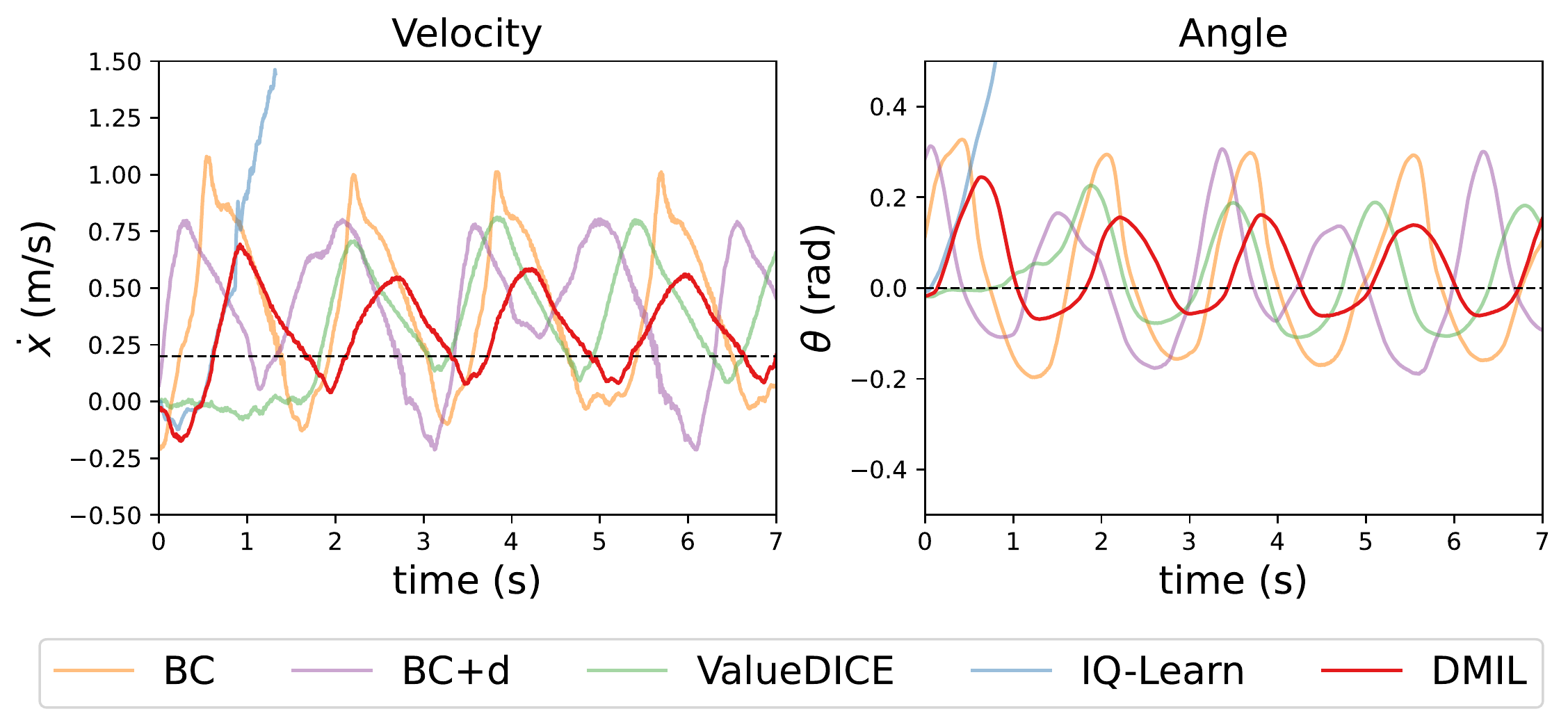}}
\end{adjustwidth}
\caption{\small Evaluation results on a real-world wheel-legged robot}
\label{fig:robot}
\makebox[\textwidth][c]{\includegraphics[width=1\textwidth]{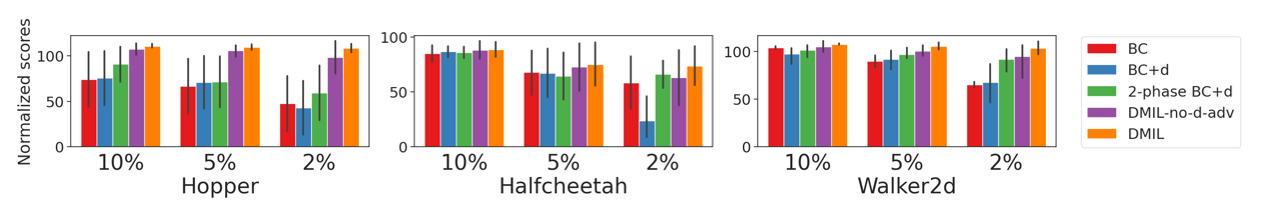}}
\caption{\small Evaluation results on policy robustness. For different sizes of expert datasets, we randomly pick 20\% samples and add a Gaussian noise on the states to make policy learning more challenging.}\label{table2}
\makebox[\textwidth][c]{\includegraphics[width=1\textwidth]{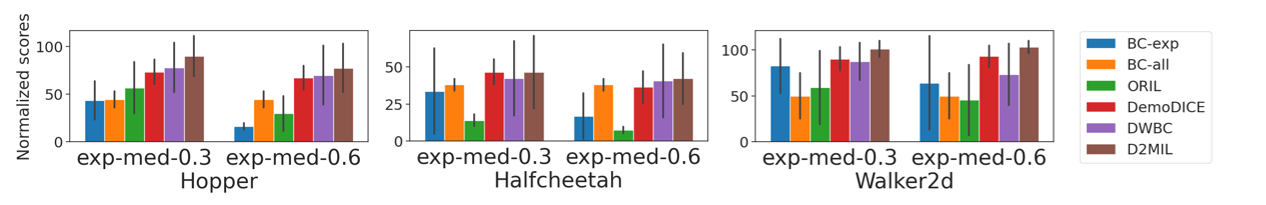}}
\caption{\small Evaluation results of D2MIL under small datasets. We first sample 1\% trajectories from the D4RL MuJoCo expert datasets. We then sample $X$ proportion of these trajectories and combine them with the 2\% medium dataset to constitute the suboptimal dataset $\mathcal{D}_o$. The remaining 1-$X$ trajectories constitute the expert dataset $\mathcal{D}_e$. The larger $X$, $\mathcal{D}_o$ contains more high quality data, but corresponds to a smaller expert dataset $\mathcal{D}_e$. 
	 	We label each task as exp-med-$X$ in the figure.}\label{table3}
\end{figure}

\textbf{Comparative Evaluation on Real-World Tasks. }
	The imitation performance of two tasks on a wheel-legged robot are shown in figure \ref{fig:robot}. 
	In these two tasks, we only use 50s human demonstrations to learn the policy.
	For the Standing still task, despite some small drifts, the robot using DMIL policy can maintain in a balanced state for over 30s, which achieves the most stable performance in both displacement measure and tilt angle. The robot with other control policies either quickly bump to the ground (BC,BC+d,valueDICE) or dashes forward (IQ-learn).
	 For the Moving straight task, most methods can make the robot move forward within a certain speed range, but DMIL policy maintains a closest speed to the target speed $v$=0.2m/s and also keeps a relatively more balanced state. 

\textbf{Evaluation on Policy Robustness. }
	We further evaluate the policy robustness of DMIL under small and noisy training data on MuJoCo tasks in Figure \ref{table2}. 
	We compare with three stronger baselines in Table~\ref{table1}: BC, BC+d and 2-phase BC+d.
	To further examine the effectiveness of the cooperative-yet-adversarial learning scheme on the learned dynamics model, we add an additional baseline DMIL-no-d-adv, which removes $\mathcal{L}_f^{corr}$ as well as $\log f$ in the input of discriminator $d$ from DMIL.
	We observe that the performances of BC and BC+d drop with the introduction of noise, mainly due to the lack of discrimination on data quality. 2-phase BC+d is slightly better, but still perform worse than DMIL and DMIL-no-d-adv. Due to the absence of adversarial learning in dynamics model, DMIL-no-d-adv is generally less performant compared with DMIL due to the noisy training data.
	In all tasks, DMIL shows great robustness to training noise and achieves almost the same performance as the case without noise (Table~\ref{table1}). This is because that the discriminator of DMIL in this setting not only distinguishes dynamics correctness and optimality of rollouts, but can also serve as a denoiser to identify and alleviate the negative impact of noisy inputs for policy and dynamics model.

\textbf{Evaluation of D2MIL. }
	We also evaluate the performance of D2MIL when learning with a small expert dataset and a larger suboptimal dataset in Figure~\ref{table3}.
	The results show that D2MIL outperforms state-of-the-art method DWBC~\cite{dwbc} and other baselines in all tasks.
	The introduction of the dynamics model $f$ and the two discriminators ($d_r$ and $d_o$) indeed help with improving the generalization performance of imitating policy under small datasets, which demonstrates the effectiveness of D2MIL in  scenarios with suboptimal data.

	\section{Related Work}
	\textbf{Model-based Imitation Learning.}
	To combat the covariate shift and improve sample efficiency, many online IL studies have incorporated dynamics models during policy learning~\cite{ode,bc3,v-mail,robotic1}. These methods typically require online system interactions or additional expert guidance to correct model errors.
	Under offline settings without environment interaction, incorporating the model-based approach is much more challenging and less explored. A few existing works all bear some limitations, such as requiring an extra suboptimal dataset~\cite{milo} or a misspecified simulator~\cite{nips2020}, only applicable to imagery input~\cite{bc2}, or simply fully trust the learned model~\cite{ode}. Many of these methods assume sufficient coverage of demonstration data, which can be fragile in scenarios with small datasets. 
	
	\textbf{Offline Imitation Learning.} Offline IL methods that imitate expert demonstrations can be categorized into two paradigms, behavior cloning (BC) and offline inverse reinforcement learning (offline IRL). 
	BC~\cite{bc} is the simplest IL method, it trains a policy by maximizing the log-likelihood of observed actions. Some recent works enhance BC 
	by using energy-based model~\cite{implicit,edm} or introducing curriculum training strategy~\cite{coil}. Offline IRL methods~\cite{valuedice,swamy2021moments,dualdice,edm,iqlearn} on the other hand, consider matching the reward or state-action distribution of the expert policy.
    This can be done explicitly by learning a reward function~\cite{swamy2021moments} or implicitly by learning a Q-function that represents both reward and policy~\cite{valuedice,iqlearn}.
	Although these recent methods can mitigate covariate shift to some extent, they still struggle to work under limited expert data and suffer from the involvement of suboptimal data.
	
	Another stream of studies focus on the problem when demonstrations contain suboptimal data. 
	Some studies~\cite{entropy,bcnd} leverage previously learned policies~\cite{bcnd} or entropy of the model~\cite{entropy} as weights to penalize noisy demonstrations. However, they require the clean expert data occupy the majority of the offline dataset.
	When both the expert demonstrations and additional suboptimal data are given, some IRL-based methods~\cite{oril,milo,demodice} first construct a reward function to distinguish expert and suboptimal data, and then use it to solve an offline RL problems. The drawbacks of these methods are that the reward learning through offline IRL is costly, and the inner-loop offline RL problem also suffers from training instability~\cite{kumar2019stabilizing}. The recently proposed DWBC~\cite{dwbc} trains a discriminator to distinguish expert and non-expert data and uses its outputs to re-weight the IL objective, so as to imitate demonstrations selectively. Our method shares some similarity with DWBC, however, we use the discriminator to distinguish both the dynamics discrepancy and suboptimality of model rollout data, and re-weight the objectives of both the IL policy and the dynamics model.

\section{Conclusion and Limitations}
We propose a model-based offline IL framework DMIL for scenarios with limited expert data, which is composed of an imitation policy, a dynamics model and a discriminator. We use the discriminator as a bridge to couple the learning process of all three models through a cooperative-yet-adversarial learning scheme.
This design allows us fully leverage the generalizability of dynamics model to improve state-action space coverage, while also alleviating the negative impacts from potentially problematic rollouts. Our framework can also be extended to scenarios with suboptimal data (D2MIL). Through comprehensive experiments, we show that our method achieves strong performance and robustness under small datasets, which can be a nice tool for many real-world IL tasks.

Our method also has some limitations. 
When the state-action space is large or the MDP is partially observed, the dynamics model might need to be specially designed. For future directions, adopting temporal models, or learning the dynamics in latent state space might be a solution to achieve improved model performance.





\acknowledgments{
We thank the anonymous reviewers for their thoughtful feedback.
This work is supported by funding from Haomo.AI.}



\bibliography{ref}

\newpage
\appendix
\section*{Appendix}
\section{Derivation Details of DMIL and D2MIL}\label{app:derivation}
	In this section, we provide the complete theoretical derivation of DMIL and D2MIL in Section~\ref{app:derivation_dmil} and \ref{app:derivation_d2mil}. As D2MIL is a direct extension of DMIL with the addition of a second optimality discriminator, hence we will only discuss the detail model design philosophy of DMIL.
	
	\subsection{Derivation Details of DMIL}\label{app:derivation_dmil}
	\textbf{A Na\"ive Model-Based Offline IL Framework. } We begin the derivation of DMIL by first inspecting the following na\"ive model-based offline IL framework, which simply incorporates a learned probabilistic dynamics model $f(s'|s,a)$ to generate rollout data $\mathcal{D}_r$ for policy learning:
	\begin{align}
		&\text{BC policy learning objective: }\quad\quad\quad\;\;\, \min\limits_{\pi} \mathcal{L}_{\pi} := \mathbb{E}_{(s,a)\sim \mathcal{D}_e}[-\log\pi(a|s)]\label{eq:app_bc} \\
		&\text{Dynamics model learning objective: }\quad \min\limits_{f} \mathcal{L}_{f} := \mathbb{E}_{(s,a,s')\sim \mathcal{D}_e}[-\log f(s'|s,a)]\label{eq:app_f} \\
		&\text{Policy learning with }\mathcal{D}_e\text{ and }\mathcal{D}_r\text{ : }\quad\quad\; \min\limits_{\pi} \mathcal{L}_{\pi}^\text{fine-tune} := \mathbb{E}_{(s,a)\sim \mathcal{D}_e \cup \mathcal{D}_r}[-\log\pi(a|s)]\label{eq:app_bc_finetune}
	\end{align}
	Specifically, when we only use Eq.(\ref{eq:app_f}) and (\ref{eq:app_bc_finetune}), it corresponds to the BC+d baseline in Section~\ref{sec:exp}; if we first use Eq.(\ref{eq:app_bc}) and (\ref{eq:app_f}) to pretrain the rollout policy and dynamics model to generate rollouts $\mathcal{D}_r$, then use Eq.(\ref{eq:app_bc_finetune}) to fine-tune the policy, this corresponds to the 2-phase-BC+d baseline.
	Obviously, these two methods all bear some drawbacks. Both methods fully trust the model rollout data, which can be problematic when the dynamics model has high prediction errors or the policy is suboptimal. Although 2-phase-BC-d uses the higher quality pretrained dynamics model and policy to generate rollouts, it may still  suffer from performance degeneration when the expert dataset is small.
	
	A remedy for this is to selectively trust and train on good rollout data, but penalize the learning on problematic rollouts. A seemingly valid approach is to jointly learn a discriminator $d(s,a)$ together with policy $\pi$ and dynamics model $f$ to judge the dynamics correctness and optimality of rollouts in a GAN-like framework~\cite{goodfellow2014generative}. In this paradigm, $\pi$ and $f$ are jointly treated as the generator and optimized implicitly through solving a min-max optimization problem on the discriminator loss $\mathcal{L}_d$, which is the cross-entropy loss between $\mathcal{D}_e$ and $\mathcal{D}_r$. Although looks reasonable, this approach faces several technical problems. First, solving the GAN-style min-max optimization problem is costly and known to suffer from training instability and issues like mode collapse~\cite{arjovsky2017wasserstein}. Second, as data in $\mathcal{D}_r$ are generated from a special multi-step rollout process using both $\pi$ and $f$, rather than single-step outputs directly from a generator model in typical GAN framework, obtaining the correct gradients of $\pi$ and $f$ for back propagation through  the discriminator loss $\mathcal{L}_d$ can be highly complex. Lastly, although we have explicit loss functions for policy $\pi$ (Eq.(\ref{eq:app_bc}) or (Eq.(\ref{eq:app_bc_finetune})) and $f$ (Eq.(\ref{eq:app_f})), they are not used to learn $\pi$ and $f$ in such a GAN-style framework. This could cause potential loss of information and performance degeneration when the expert data $\mathcal{D}_e$ contain noisy or suboptimal data. Since under the GAN framework, the only objectives of $\pi$ and $f$ are to fool the discriminator, rather than maximizing the likelihood on expert data.
	
	
	\textbf{Problem Reformulation Under the Cooperative-yet-Adversarial Learning Scheme. } To address above issues, we introduce an  adversarial-yet-cooperative learning scheme to jointly learn the policy $\pi$, dynamics model $f$ and discriminator $d$. In particular, we first include the element-wise loss information from policy and dynamics model ($\log\pi$ and $\log f$) into the inputs of the discriminator $d$ (i.e., $d(s,a,\log\pi(a|s),\log f(s'|s,a))$) to establish cooperative information sharing, and then use the following adversarial learning objective to learn the discriminator $d$:
	\begin{equation}
		\begin{aligned}
			\underset{d}{\min}\, \underset{\pi,f}{\max}\; \mathcal{L}_d := &\underset{(s,a,s')\sim\mathcal{D}_{e}}{\mathbb{E}}[-\log d(s,a,\log\pi(a|s),\log f(s'|s,a))]+\\
			&\underset{(s,a,s')\sim\mathcal{D}_{r}}{\mathbb{E}}[-\log(1-d(s,a,\log\pi(a|s),\log f(s'|s,a)))] \label{eq:app_d}
		\end{aligned}
	\end{equation}
	Although this design looks not very intuitive, we can show that it offers a series of benefits. First, the information sharing couples the learning process of $\pi$, $f$ and $d$, and also provides valuable information for $d$ to make better judgment, as discussed in the main article in Section~\ref{sec:method_dmil}. Second, making $\pi$ and $f$ challenge the discriminator $d$ by injecting adversarial information through $\log\pi(a|s)$ and $\log f(s'|s,a)$ will force the discriminator $d$ to minimize the worst-case error of $\mathcal{L}_d$, which has been shown in adversarial learning studies to greatly improve model robustness~\cite{carlini2019evaluating,goodfellow2014explaining}. Last and most importantly, we can show that this design enables reformulating the original complex coupled optimization problems (LHS of Eq.(\ref{eq:app_formulation})) into three simple minimization problems as follows, which can be easily solved in a fully supervised learning manner to achieve high computation efficiency.
	\begin{equation}
		\left\{\begin{array}{l}
			\min_{\pi}\; \mathcal{L}_{\pi} \\
			\min_{f}\; \mathcal{L}_{f} \\
			\underset{d}{\min}\, \underset{\pi,f}{\max} \; \mathcal{L}_d
		\end{array}\right.
		\quad\Rightarrow\quad
		\left\{\begin{array}{l}
			\min_{\pi}\; \mathcal{L}_{\pi}^\text{DMIL} := \alpha_{\pi}\cdot\mathcal{L}_{\pi}+\mathcal{L}_{\pi}^{corr} \\
			\min_{f}\; \mathcal{L}_{f}^\text{DMIL} := \alpha_f\cdot\mathcal{L}_{f}+\mathcal{L}_{f}^{corr} \\
			\min_{d}\; \mathcal{L}_d
		\end{array}\right.\label{eq:app_formulation}
	\end{equation}
	where $\mathcal{L}_\pi$ and $\mathcal{L}_f$ are defined on $\mathcal{D}_e$ as shown in Eq.(\ref{eq:app_bc}) and (\ref{eq:app_f}); $\mathcal{L}_{\pi}^{corr}$ and $\mathcal{L}_{f}^{corr}$ are corrective loss terms capturing the adversarial behavior of $\pi$ and $f$ on $d$, which are computed based on output values of the discriminator $d$ on samples from both $\mathcal{D}_e$ and $\mathcal{D}_r$; $\alpha_{\pi}$, $\alpha_f \geq 1$ are weight factors of $\pi$ and $f$ to balance their original learning objectives and the additional adversarial behavior.
	
	
	The corrective loss terms $\mathcal{L}_{\pi}^{corr}$ and $\mathcal{L}_{f}^{corr}$ are derived by finding equivalent relaxed conditions of the inner maximization problem for $\pi$ and $f$ in $\underset{d}{\min}\, \underset{\pi,f}{\max} \,\mathcal{L}_d$. This avoids solving the original complex functional min-max problem for the discriminator, and also enables learning $\pi$ and $f$ on both expert data $\mathcal{D}_e$ and model rollouts $\mathcal{D}_r$. Utilizing calculus of variation~\cite{gelfand2000calculus} and the analysis method introduced in \citet{dwbc}, we provide the detailed derivation of the exact forms of $\mathcal{L}_{\pi}^{corr}$ and $\mathcal{L}_{f}^{corr}$ as follows.
	

	\textbf{Derivation of the Corrective Loss Terms. } Under the proposed cooperative-yet-adversarial learning scheme, both the discriminator $d$ and its loss $\mathcal{L}_d$ become functionals of $\pi$ and $f$ (i.e., function of a function), which can be expressed as $d(s,a,\log\pi(a|s),\log f(s'|s,a))$ and $\mathcal{L}_d(d,\log\pi, \log f)$. 
	Denote $x=(s,a,s')$. Note that $\mathcal{L}_d(d,\log\pi, \log f)$ can be rewritten as following integral form of a new functional $F(x,d,\log\pi, \log f)$:
	\begin{align}
		\mathcal{L}_d(d,\log\pi, \log f)=&\underset{(s,a,s')\sim\mathcal{D}_{e}}{\mathbb{E}}[-\log d(s,a,\log\pi(a|s),\log f(s'|s,a))]\notag\\
		&+\underset{(s,a,s')\sim\mathcal{D}_{r}}{\mathbb{E}}[-\log(1-d(s,a,\log\pi(a|s),\log f(s'|s,a)))]\notag\\
		=&\int_{\Omega_{sas'}}\big[P_{D_e}(x)\cdot[-\log d(s,a,\log\pi(a|s),\log f(s'|s,a))]\notag\\
		&+P_{D_r}(x)\cdot[-(1-\log d(s,a,\log\pi(a|s),\log f(s'|s,a)))]\big]\mathrm{d}x \notag\\
		\triangleq& \int_{\Omega_{sas'}} F(x,d,\log\pi, \log f) \mathrm{d}x
	\end{align}
	where $P_{D_e}$ and $P_{D_r}$ are probability distributions of $x$ in $\mathcal{D}_e$ and $\mathcal{D}_r$; and $\Omega_{sas'}$ is the domain of $x$ under $\mathcal{D}_e \cup \mathcal{D}_r$.
	
	To avoid solving the complex functional min-max problem $\underset{d}{\min}\, \underset{\pi,f}{\max}\, \mathcal{L}_d(d,\log\pi, \log f)$, we will focus on its inner maximization problem, which essentially requires to find the maxima of functional $\mathcal{L}_d(d,\log\pi, \log f)$ with respect to $\pi$ and $f$, given an unknown functional $d$ decided by the outer minimization problem.
	From functional analysis and calculus of variation\cite{gelfand2000calculus}, 
	the extrema (maxima or minima) of $\mathcal{L}_d$ can be obtained by solving the following associate Euler-Lagrangian equations:
	\begin{equation}
		\left\{\begin{array}{l}
			F_{\pi}-\frac{\partial}{\partial x} F_{\frac{\partial \pi}{\partial x}}=F_{\pi}=0 \\
			F_{f}-\frac{\partial}{\partial x} F_{\frac{\partial f}{\partial x}}=F_{f}=0
		\end{array}\right.
	\end{equation}

	where $F_y$ stands for $\frac{\partial F}{\partial y}$. As $\frac{\partial \pi}{\partial x}$, and $\frac{\partial f}{\partial x}$ do not appear in the our form of $F(x,d,\log\pi, \log f)$, hence $F_{\frac{\partial \pi}{\partial x}}=F_{\frac{\partial f}{\partial x}}=0$. Let $\theta_\pi$ and $\theta_f$ denote model parameters of $\pi$ and $f$, above equations also indicate:
	\begin{equation}
		\left\{\begin{array}{l}
			F_{\pi}\cdot \frac{\partial \pi}{\partial \theta_{\pi}}=\frac{\partial F}{\partial d} \cdot \frac{\partial d}{\partial \log\pi} \cdot \frac{\partial \log\pi}{\pi}\cdot \frac{\partial \pi}{\partial \theta_{\pi}}=\frac{\partial F}{\partial d} \cdot \frac{\partial d}{\partial \log\pi}\cdot\nabla_{\theta_{\pi}}\log\pi=0 \\
			F_{f}\cdot \frac{\partial f}{\partial \theta_{f}}=\frac{\partial F}{\partial d} \cdot \frac{\partial d}{\partial \log f} \cdot \frac{\partial \log f}{f}\cdot \frac{\partial f}{\partial \theta_{f}}=\frac{\partial F}{\partial d} \cdot \frac{\partial d}{\partial \log f}\cdot\nabla_{\theta_{f}}\log f=0
		\end{array}\right.
	\end{equation}
	
	
	
	In our problem, $d$, $F$, $\pi$ and $f$ are real-value functions, hence the same with the derivatives $\frac{\partial F}{\partial d}$, $\frac{\partial d}{\partial \log\pi}$ and $\frac{\partial d}{\partial \log f}$. If the continuity of previous functions and derivatives are satisfied, then according to \citet{hewitt1948rings}, the set of real-valued continuous functions is a commutative ring, we can safely swap the order of $\frac{\partial F}{\partial d}$ and $\frac{\partial d}{\partial \log\pi}$, as well as $\frac{\partial F}{\partial d}$ and $\frac{\partial d}{\partial \log f}$ in above equations.
	
	As $d$ is determined by the outer minimization problem of Eq.(\ref{eq:app_d}), thus the exact forms of $\frac{\partial d}{\partial \log\pi}$ and $\frac{\partial d}{\partial \log f}$ are not obtainable by only inspecting the inner maximization problem. We can instead consider a alternative solution by making $\frac{\partial F}{\partial d}\cdot\nabla_{\theta_{\pi}}\log\pi=0$ and $\frac{\partial F}{\partial d}\cdot\nabla_{\theta_{f}}\log f=0$ for state-action pairs in $\Omega_{s}\times\Omega_{a}$. For practical IL tasks, $\mathcal{D}_e$ and $\mathcal{D}_r$ are finite, and the domains $\Omega_{s}$ and $\Omega_{a}$ are closed and bounded, hence the integration on $\frac{\partial F}{\partial d}\cdot\nabla_{\theta_{\pi}}\log\pi$ and  $\frac{\partial F}{\partial d}\cdot\nabla_{\theta_{f}}\log f$ will still be zero. 
	Interestingly, although it is intractable to directly solve $\frac{\partial F}{\partial d}\cdot\nabla_{\theta_{\pi}}\log\pi=0$ and $\frac{\partial F}{\partial d}\cdot\nabla_{\theta_{f}}\log f=0$, the integration on these equations leads to two new relaxed and tractable necessary conditions for $\mathcal{L}_d$ to reach its extrema. Using the condition on $\pi$ as an example, we have:
	\begin{align}
		0=&\int_{\Omega_{sas'}} \frac{\partial F(x, d, \pi(a|s),f(s'|s,a))}{\partial d(s, a,\pi(a|s),f(s'|s,a))} \cdot \nabla_{\theta_{\pi}}\log\pi(a|s) \mathrm{d}x \notag\\
		=&\int_{\Omega_{sas'}} 
		\bigg[-P_{\mathcal{D}_{e}}(x) \cdot\frac{1}{d(s, a, \log\pi(a|s),\log f(s'|s,a))}\notag\\
		&+P_{\mathcal{D}_{o}}
		(x) \cdot\frac{1}{1-d(s, a,
			\log\pi(a|s),\log f(s'|s,a))}\bigg]
		\cdot \nabla_{\theta_{\pi}}\log\pi(a|s) \mathrm{d}x \notag\\
		=&\underset{(s,a,s')\sim\mathcal{D}_e}{\mathbb{E}}\left[-\frac{1}{d}\cdot\nabla_{\theta_{\pi}}\log\pi\right]- \underset{(s,a,s')\sim\mathcal{D}_r}{\mathbb{E}}\left[-\frac{1}{1-d}\cdot\nabla_{\theta_{\pi}}\log\pi\right]\label{eq:app_equi_cond}
	\end{align}
	where in the last equation, we slightly abuse the notations and write the output value of $d(s,a,\log\pi(a|s),\log f(s'|s,a))$ as $d$.
	%
	Note that the above condition can be equivalently perceived as the first-order optimality condition of minimizing a new loss term $\mathcal{L}_{\pi}^{corr}$ with respect to $\pi$, i.e., derivative equal to zero, given as
	\begin{equation}
		\mathcal{L}_{\pi}^{corr}=-\underset{(s,a,s')\sim\mathcal{D}_e}{\mathbb{E}} \left[-\frac{1}{d}\cdot\log\pi(a|s)\right]+ \underset{(s,a,s')\sim\mathcal{D}_r}{\mathbb{E}} \left[-\frac{1}{1-d}\cdot\log\pi(a|s)\right]\label{eq:app_corr_pi}
	\end{equation}
	where we introduce a negative sign on the last equation in Eq.(\ref{eq:app_equi_cond}) to ensure minimizing $\mathcal{L}_{\pi}^{corr}$ leads to update $\pi$ in the gradient ascent direction of $\mathcal{L}_d$, so as to find the maxima of $\mathcal{L}_d$ rather than minima. 
	
	Similarly to the derivation of $\mathcal{L}_{\pi}^{corr}$, we can get the corrective loss for the dynamics model $\mathcal{L}_{f}^{corr}$ as:
	\begin{equation}
		\mathcal{L}_{f}^{corr}=-\underset{(s,a,s')\sim\mathcal{D}_e}{\mathbb{E}} \left[-\frac{1}{d}\cdot\log f(s'|s,a)\right]+ \underset{(s,a,s')\sim\mathcal{D}_r}{\mathbb{E}} \left[-\frac{1}{1-d}\cdot\log f(s'|s,a)\right]\label{eq:app_corr_f}
	\end{equation}
	
	Add these corrective loss terms to their original losses according to Eq.(\ref{eq:app_formulation}), we can get the final objectives for $\pi$ and $f$ in DMIL:
	\begin{align}
		\mathcal{L}^\text{DMIL}_{\pi}=&\alpha_{\pi}\underset{(s,a)\sim\mathcal{D}_e}{\mathbb{E}}\left[-\log\pi(a|s)\right]-\underset{(s,a,s')\sim\mathcal{D}_e}{\mathbb{E}}\left[-\frac{1}{d}\cdot\log\pi(a|s)\right]+ \underset{(s,a,s')\sim\mathcal{D}_r}{\mathbb{E}}\left[-\frac{1}{1-d}\cdot\log\pi(a|s)\right] \notag\\
		=&\underset{(s,a,s')\sim\mathcal{D}_e}{\mathbb{E}}\left[-\left(\alpha_{\pi}-\frac{1}{d}\right)\cdot \log\pi(a|s)\right]+\underset{(s,a,s')\sim\mathcal{D}_r}{\mathbb{E}}\left[-\frac{1}{1-d}\cdot\log\pi(a|s)\right] \label{eq:app_dmil_final_pi}
	\end{align}
	\begin{align}
		\mathcal{L}^\text{DMIL}_{f}= &
		\alpha_f \underset{(s,a,s')\sim\mathcal{D}_e}{\mathbb{E}}\left[-\log f(s'|s,a)\right]-\underset{(s,a,s')\sim\mathcal{D}_e}{\mathbb{E}}\left[-\frac{1}{d}\cdot\log f(s'|s,a)\right]+ \underset{(s,a,s')\sim\mathcal{D}_r}{\mathbb{E}}\left[-\frac{1}{1-d}\cdot\log f(s'|s,a)\right] \notag\\
		=&\underset{(s,a,s')\sim\mathcal{D}_e}{\mathbb{E}}\left[-\left(\alpha_{f}-\frac{1}{d}\right)\cdot \log f(s'|s,a)\right]+\underset{(s,a,s')\sim\mathcal{D}_r}{\mathbb{E}}\left[-\frac{1}{1-d}\cdot\log f(s'|s,a)\right]
		\label{eq:app_dmil_final_f}
	\end{align}
	
	Note that we use $d(s,a,\log\pi(a|s),\log f(s'|s,a))$ as values in $\mathcal{L}_{\pi}^{corr}$ and $\mathcal{L}_{f}^{corr}$, thus there is no gradient passing from the discriminator $d$ to $\pi$ and $f$ when minimizing $\mathcal{L}^\text{DMIL}_{\pi}$ and $\mathcal{L}^\text{DMIL}_{f}$. This greatly simplifies the learning processes of $\pi$, $f$ and $d$, as all of them can be trained in a decoupled manner with their own optimization objectives (Eq.(\ref{eq:app_formulation})), while also enabling capturing the coupled relationship with $d$ using $\mathcal{L}_{\pi}^{corr}$ and $\mathcal{L}_{f}^{corr}$.

	
	\textbf{Interpretations of DMIL. } The final learning objectives of $\pi$ and $f$ in Eq.(\ref{eq:app_dmil_final_pi}) and (\ref{eq:app_dmil_final_f}) are actually intuitively reasonable. It can be perceived as assigning credibility weights on different samples based on the judgment of the discriminator $d$, with weight $\alpha_\pi-1/d$ and $\alpha_f-1/d$ assigned to expert demonstrations and $1/(1-d)$ assigned to model rollout data. Suppose the discriminator is well-learned, then it will output small values for problematic model rollouts, resulting in lower weights ($1/(1-d)\rightarrow 1$) on these samples; whereas for credible rollout samples ($d\rightarrow 1$), the weights will be boosted and encourage the policy $\pi$ to learn more on these samples. Moreover, the learned discriminator can also serve as a denoiser to alleviate noisy or suboptimal data in the expert dataset $\mathcal{D}_e$. For such samples, the output values of $d$ will be small, and the weights $\alpha_\pi-1/d$ and $\alpha_f-1/d$ will be reduced for policy $\pi$ and dynamics model $f$.
	
	It should be noted that during our derivation, the continuity assumption of $\frac{\partial F}{\partial d}$ needs to be satisfied. We thus clip the output range of $d$ to $[0.1,0.9]$ to avoid $1/d$ and $1/(1-d)$ taking infinite values. We further set $\alpha_{\pi}=\alpha_{f}=10$ in our implementation to ensure expert demonstrations in $\mathcal{D}_e$ always get positive weights.
	
	
	\subsection{Derivation Details of D2MIL}\label{app:derivation_d2mil}
	\textbf{Problem Formulation of D2MIL. }
	As for offline IL scenarios with a small expert dataset $\mathcal{D}_{e}$ and a large unknown, potentially suboptimal dataset $\mathcal{D}_{o}$, we can extend the proposed DMIL framework by adding a second optimality discriminator $d_o(s,a,\log\pi)$ to distinguish expert and non-expert samples, following a similar treatment as in DWBC~\cite{dwbc}. Moreover, we also introduce a second pair of adversarial relationship between the policy $\pi$ and $d_o$ to carry over the similar reformulation design as in DMIL. For clarity, we will refer the original rollout discriminator in DMIL as $d_r$ in the following discussion. Under this scenario, the set of problems we need to jointly solve are:
	\begin{equation}
		\left\{\begin{array}{l}
			\min_{\pi}\; \mathcal{L}_{\pi}:= \mathbb{E}_{(s,a)\sim \mathcal{D}_e}[-\log \pi(a|s)]\\
			\min_{f}\; \mathcal{L}^{'}_{f}:= \mathbb{E}_{(s,a,s')\sim \mathcal{D}_e\cup\mathcal{D}_o}[-\log f(s'|s,a)] \\
			\underset{d_r}{\min}\, \underset{\pi,f}{\max} \; \mathcal{L}_{d_r}\\
			\underset{d_o}{\min}\, \underset{\pi}{\max} \; \mathcal{L}_{d_o}\\
		\end{array}\right.\label{eq:app_d2mil_formulation}
	\end{equation}
	where we use the same policy learning objective $\mathcal{L}_{\pi}$ to make it only learn from the expert demonstrations, but use an updated objective $\mathcal{L}^{'}_{f}$ for the dynamics model $f$, as it can learn from both the real expert and suboptimal datasets $\mathcal{D}_e\cup\mathcal{D}_o$ regardless of the optimality of data. For the rollout discriminator $d_r$, now it needs to distinguish both the real expert and suboptimal data $\mathcal{D}_e\cup\mathcal{D}_o$ from model generated rollouts $\mathcal{D}_r$, hence we update its learning objective as follows:
	\begin{equation}
		\begin{aligned}
			\mathcal{L}_{d_r} = &\underset{(s,a,s')\sim\mathcal{D}_{e}\cup\mathcal{D}_o}{\mathbb{E}}[-\log d(s,a,\log\pi(a|s),\log f(s'|s,a))]+\\
			&\underset{(s,a,s')\sim\mathcal{D}_{r}}{\mathbb{E}}[-\log(1-d(s,a,\log\pi(a|s),\log f(s'|s,a)))] \label{eq:app_d2mil_d}
		\end{aligned}
	\end{equation}
	For the additional optimality discriminator $d_o$, we follow the treatment in previous works~\cite{oril,dwbc} to adopt a positive-unlabeled (PU) learning~\cite{pu-learn} objective, as the the unknown suboptimal dataset $\mathcal{D}_{o}$ may also contain some expert-like data. Utilizing PU learning allows us to learn from positive (expert data $\mathcal{D}_{e}$) and unlabeled data ($\mathcal{D}_{o}\cup\mathcal{D}_{r}$ in our case). The learning objective of $d_o$ is given as:
	\begin{align}
			\mathcal{L}_{d_o} = &\eta \underset{(s,a)\sim\mathcal{D}_{e}}{\mathbb{E}}[-\log d_o(s,a,\log\pi(a|s))]+
			\underset{(s,a)\sim\mathcal{D}_{o}\cup\mathcal{D}_{r}}{\mathbb{E}}[-\log(1-d_o(s,a,\log\pi(a|s)))]\notag\\
			&-
			\eta\underset{(s,a)\sim\mathcal{D}_{e}}{\mathbb{E}}[-\log(1-d_o(s,a,\log\pi(a|s)))]
		\label{eq:app_do}
	\end{align}
	where $\eta$ is a hyperparameter, corresponds to the proportion of positive samples to unlabeled samples. We set it as 0.5 in all our experiments.
	
	Following a similar reformulation scheme as in DMIL, we can avoid solving the two complex functional min-max optimization problems in Eq.(\ref{eq:app_d2mil_formulation}) by considering the following reformulation:
	
	\begin{equation}
		\left\{\begin{array}{l}
			\min_{\pi}\; \mathcal{L}_{\pi}^\text{D2MIL} := \alpha_{\pi}\cdot\mathcal{L}_{\pi}+\mathcal{L}_{\pi}^{corr} = \alpha_{\pi}\cdot\mathcal{L}_{\pi}+\beta_r\cdot\mathcal{L}_{\pi}^{corr_r} + \beta_o\cdot\mathcal{L}_{\pi}^{corr_o} \\
			\min_{f}\; \mathcal{L}_{f}^\text{D2MIL} := \alpha_f\cdot\mathcal{L}_{f}^{'}+\mathcal{L}_{f}^{corr} \\
			\min_{d_r}\; \mathcal{L}_{d_r}\\
			\min_{d_o}\; \mathcal{L}_{d_o}
		\end{array}\right.\label{eq:app_d2mil_reformulation}
	\end{equation}
	Due to the existence of two pairs of adversarial relationships involving policy $\pi$, the corrective loss term on $\pi$ will become the sum of two terms, i.e., $\mathcal{L}_{\pi}^{corr}=\beta_r\cdot\mathcal{L}_{\pi}^{corr_r} + \beta_o\cdot\mathcal{L}_{\pi}^{corr_o}$. $\beta_r$ and $\beta_o$ are the weight factors to balance the impact from  both the original rollout discriminator $d_r$ and the optimality discriminator $d_o$ on policy $\pi$. To reduce the number of hyperparameters in the model, we set $\beta_o=1-\beta_r$. The derivation of the exact forms of $\mathcal{L}_{\pi}^{corr_r}$, $\mathcal{L}_{\pi}^{corr_o}$ and $\mathcal{L}_f$ under D2MIL are described below.

	\textbf{Corrective Loss Terms under D2MIL. } Following the same derivation procedure of DMIL in Appendix~\ref{app:derivation_dmil}, the updated corrective loss terms $\mathcal{L}_f^{corr}$ and $\mathcal{L}_{\pi}^{corr_r}$ for dynamics model $f$ and policy $\pi$ under D2MIL can be easily obtained as follows:
	\begin{align}
			\mathcal{L}_f^{corr}&=
			-\underset{(s,a,s')\sim\mathcal{D}_e\cup\mathcal{D}_o}{\mathbb{E}}\left[-\frac{1}{d_r}\cdot\log f(s'|s,a)\right]+ \underset{(s,a,s')\sim\mathcal{D}_r}{\mathbb{E}}\left[-\frac{1}{1-d_r}\cdot\log f(s'|s,a)\right]
			\label{eq:app_d2mil_corr_f} \\
			\mathcal{L}_{\pi}^{corr_r}&=
			-\underset{(s,a,s')\sim\mathcal{D}_e\cup\mathcal{D}_o}{\mathbb{E}}\left[-\frac{1}{d_r}\cdot\log \pi(a|s)\right]+ \underset{(s,a,s')\sim\mathcal{D}_r}{\mathbb{E}}\left[-\frac{1}{1-d_r}\cdot\log \pi(a|s)\right]
			\label{eq:app_d2mil_corr_pi_r}
	\end{align}

	While for the learning objective of discriminator $d_o$ in Eq.(\ref{eq:app_do}), let $z=(s,a)$ and $\Omega_{sa}$ as its domain, then it can be rewritten as the integral of a new functional $F_o(z,d_o,\log\pi(a|s))$:
	\begin{align}
			\mathcal{L}_{d_o}=&\int_{\Omega_{sa}}\Big[
			P_{D_e}(z)\cdot\eta[-\log d_o(z,\log\pi(a|s))]+(P_{D_o}(z)+P_{D_r}(z))\cdot[-\log(1- d_o(z,\log\pi(a|s)))]\notag\\
			&-
			P_{D_e}(z)\cdot\eta[-\log(1- d_o(z,\log\pi(a|s)))]\Big]\mathrm{d}z \notag\\
			\triangleq& \int_{\Omega_{sa}} F_o(z,d_o,\log\pi(a|s)) \mathrm{d}z
	\end{align}
	where $P_{D_e}(z)$, $P_{D_o}(z)$ and $P_{D_r}(z)$ are the probability distributions of $z$ in $\mathcal{D}_e$, $\mathcal{D}_o$ and $\mathcal{D}_r$, respectively.
	Following the derivation in previous section, we can get the similar relaxed necessary condition for $\mathcal{L}_{d_o}$ to reach its extrema with respect to $\pi$ as:
	\begin{align}
			&\int_{\Omega_{sa}} \frac{\partial F_o(z, d_o, \log\pi(a|s))}{\partial d_o(z,\log\pi(a|s))} \cdot \nabla_{\theta_{\pi}}\log\pi(a|s) \mathrm{d}z \notag\\
			=&\int_{\Omega_{sa}}
			\bigg[-P_{\mathcal{D}_{e}}(z) \cdot\frac{\eta}{d_o(z, \log\pi(a|s))}
			+(P_{D_o}(z)+P_{D_r}(z))\cdot\frac{1}{1-d_o(z,
				\log\pi(a|s))}\notag\\
			&-P_{\mathcal{D}_{e}}(z) \cdot\frac{\eta}{1-d_o(z, \log\pi(a|s))} \bigg]
			\cdot \nabla_{\theta_{\pi}}\log\pi(a|s) \mathrm{d}z\notag\\
			=&\underset{(s,a)\sim\mathcal{D}_e}{\mathbb{E}}\left[-\frac{\eta}{d_o}\cdot\nabla_{\theta_{\pi}}\log\pi(a|s)\right]- \underset{(s,a)\sim\mathcal{D}_o\cup\mathcal{D}_r}{\mathbb{E}}\left[-\frac{1}{1-d_o}\cdot\nabla_{\theta_{\pi}}\log\pi(a|s)\right]\notag\\
			&+\underset{(s,a)\sim\mathcal{D}_e}{\mathbb{E}}\left[-\frac{\eta}{1-d_o}\cdot\nabla_{\theta_{\pi}}\log\pi(a|s)\right]=0\label{eq:app_d2mil_condition}
	\end{align}
	Again, we slightly abuse the notations and write the output values of $d_o(s,a,\log(a|s))$ as $d_o$ in the last equation. Similar to the derivation of DMIL, above condition can be perceived as the first-order optimality condition of the corrective loss term $\mathcal{L}_{\pi}^{corr_o}$ with the following form:
	\begin{equation}
		\mathcal{L}_{\pi}^{corr_o}=-\underset{(s,a)\sim\mathcal{D}_e}{\mathbb{E}}\left[-\frac{\eta}{d_o(1-d_o)}\cdot\log\pi(a|s)\right]+ \underset{(s,a)\sim\mathcal{D}_o\cup\mathcal{D}_r}{\mathbb{E}}\left[-\frac{1}{1-d_o}\cdot\log\pi(a|s)\right]
		\label{eq:app_d2mil_pi_corr_o}
	\end{equation}
	Plug these corrective loss terms back to the reformulated problem in Eq.(\ref{eq:app_d2mil_reformulation}), we obtain the final learning objectives of $\pi$ and $f$ in D2MIL:
	\begin{align}
			\mathcal{L}^\text{D2MIL}_{\pi}
			=&\underset{(s, a, s') \sim \mathcal{D}_e}{\mathbb{E}}\left[-\left(\alpha_{\pi}-\frac{\beta_{o} \eta}{d_{o}\left(1-d_{o}\right)}-\frac{\beta_{r}}{d_{r}}\right)\cdot \log\pi(a|s) \right]+\underset{(s, a, s') \sim \mathcal{D}_o}{\mathbb{E}}\left[-\left(\frac{\beta_{o} }{1-d_{o}}-\frac{\beta_{r}}{d_{r}}\right)\cdot \log\pi(a|s) \right] \notag\\
			&+\underset{(s, a, s') \sim \mathcal{D}_r}{\mathbb{E}}\left[-\left(\frac{\beta_{o}}{1-d_{o}}+\frac{\beta_{r} }{1-d_{r}}\right)\cdot \log\pi(a|s) \right] \label{eq:app_d2mil_final_pi}
	\end{align}
	\begin{align}
		\mathcal{L}^\text{D2MIL}_{f}
		=\underset{(s,a,s')\sim\mathcal{D}_e\cup\mathcal{D}_o}{\mathbb{E}}\left[-\left(\alpha_{f}-\frac{1}{d_r}\right)\cdot \log f(s'|s,a)\right]+\underset{(s,a,s')\sim\mathcal{D}_r}{\mathbb{E}}\left[-\frac{1}{1-d_r}\cdot\log f(s'|s,a)\right]
		\label{eq:app_d2mil_final_f}
	\end{align}
	Again, to ensure the continuity assumption is satisfied during derivation, we clip the output range of both $d_o$ and $d_r$ to $[0.1,0.9]$.
	
	In the final objective of $\mathcal{L}^\text{D2MIL}_{\pi}$, $\beta_o$ and $\beta_r$ ($\beta_o+\beta_r=1$) actually reflect the trade-off between the reliability and optimality of samples in $\mathcal{D}_o$ and $\mathcal{D}_r$. When $\beta_o=\beta_r$, D2MIL tends to learn policy with high $d_o$ and $d_r$ samples with similar preference. However, if the suboptimal dataset $\mathcal{D}_o$ is known to have high quality, one can use a larger $\beta_r$ to pay more attention to the quality of rollout data. In such cases, both $d_o$ and $d_r$ will output values close to 1 on $\mathcal{D}_o$ samples, resulting high weights to encourage policy learning on these samples. Conversely, if the expert demonstrations $\mathcal{D}_e$ and suboptimal dataset $\mathcal{D}_o$ has considerably large gap, a large $\beta_o$ should be used to ensure policy learning focus more on those expert-like samples.
	

\section{Algorithm and Implementation Details}\label{app:details}
\subsection{Algorithm Details}\label{app:alg_details}
We outline the pseudocode of DMIL in Algorithm \ref{alg:A} and D2MIL in Algorithm \ref{alg:d2mil}. 
\begin{algorithm}
\caption{Discriminator-guided Model-based Offline Imitation Learning (DMIL)}
\label{alg:A}
\begin{algorithmic}[1]
\REQUIRE {Expert dataset $D_e$, hyperparameter $\alpha_\pi$, $\alpha_f$}
\STATE {Initialize the discriminator $d$, dynamics model $f$ and imitation policy $\pi$; set $\mathcal{D}_r=\emptyset$.}
\STATE {Train a preliminary dynamics model $f$ using samples from $D_{e}$}
\FOR{training step $t={1\cdots N}$}
    \STATE {Utilize dynamics model $f$ and imitation policy $\pi$ to generate rollouts and add into $D_r$}
    \STATE {Sample $(s_e,a_e,s'_e)\sim D_{e}$ and $(s_r,a_r,s'_r)\sim D_{r}$ to form a training batch}
    \STATE Update $d$ by minimizing the objective in Eq.(\ref{eq:app_d})
    \STATE Update $\pi$ by minimizing the objective in Eq.(\ref{eq:app_dmil_final_pi})
    \STATE Update $f$ by minimizing the objective in Eq.(\ref{eq:app_dmil_final_f})
\ENDFOR
\end{algorithmic}
\end{algorithm}

\begin{algorithm}
\caption{Dual-Discriminator Guided Model-based Offline Imitation Learning (D2MIL)}\label{alg:d2mil}
\begin{algorithmic}[1]
\REQUIRE {Expert dataset $D_e$, suboptimal dataset $D_o$, hyperparameter $\alpha_\pi$, $\alpha_f$, $\beta_r$, $\beta_o$}
\STATE {Initialize the discriminators $d_o$, $d_r$, dynamics model $f$ and imitation policy $\pi$; set $\mathcal{D}_r=\emptyset$.}
\STATE {Train a preliminary dynamics model $f$ using samples from $D_{e}\cup D_{o}$}
\FOR{training step $t={1\cdots N}$}
    \STATE {Utilize dynamics model $f$ and imitation policy $\pi$ to generate rollouts and add into $D_r$}
    \STATE {Sample $(s_e,a_e,s'_e)\sim D_{e}$, $(s_o,a_o,s'_o)\sim D_{o}$ and $(s_r,a_r,s'_r)\sim D_{r}$ to form a training batch}
    \STATE Update $d_r$ by minimizing the objective in Eq.(\ref{eq:app_d2mil_d})
    \STATE Update $d_o$ by minimizing the objective in Eq.(\ref{eq:app_do})
    \STATE Update $\pi$ by minimizing the objective in Eq.(\ref{eq:app_d2mil_final_pi})
    \STATE Update $f$ by minimizing the objective in Eq.(\ref{eq:app_d2mil_final_f})
\ENDFOR
\end{algorithmic}
\end{algorithm}

\begin{minipage}[t]{0.55\linewidth}
\centering
\small
\begin{tabular}{lcc}
\toprule
\multirow{2}{*}{Hyperparameters} & \multicolumn{2}{c}{Values in experiments} \\
                                 & D4RL tasks       & Real-world tasks       \\ \midrule
DMIL-$\alpha_\pi$  & 10               & 10                \\
DMIL-$\alpha_f$    & 10               & 10                \\\midrule
D2MIL-$\alpha_\pi$ & 10               & 10                \\
D2MIL-$\alpha_f$   & 10               & 10                \\
D2MIL-$\eta$       & 0.5              & 0.5               \\
D2MIL-$\beta_o$    & 0.5              & 0.6               \\
D2MIL-$\beta_r$    & 0.5             & 0.4            \\ \bottomrule
\end{tabular}
\captionof{table}{Hyperparemeter values.}
\label{param}
\end{minipage}
\begin{minipage}[t]{0.45\linewidth}
\centering
\small
\begin{tabular}{lc}
\toprule
Tasks &Transitions \\ \midrule
MuJoCo-exp-10\%  & 100,000        \\ 
MuJoCo-exp-5\%   & 50,000       \\ 
MuJoCo-exp-2\%   & 20,000     \\ 
Pen-human & 5,000    \\ 
Hammer-human & 11,310              \\ 
Door-human   & 6,729          \\ \midrule
exp-med-0.3         & $\mathcal{D}_e$: 7,000, $\mathcal{D}_o$: 23,000 \\ 
exp-med-0.6         & $\mathcal{D}_e$: 4,000, $\mathcal{D}_o$: 26,000 \\ \bottomrule
\end{tabular}
\captionof{table}{Datesets details for D4RL tasks.}\label{datasets}
\end{minipage}

\subsection{Implementation Details}\label{app:imp_details}
For all experiments on MuJoCo tasks, all models (dynamics model $f$, imitation policy $\pi$, discriminator $d$ ($d_r$, $d_o$ for D2MIL)) are implemented as 2-layer neural networks with 256 hidden units each layer for dynamics model and policy, and 512 hidden units for the discriminator. For Adroit tasks, we use the same network configuration for dynamics model and discriminator, but increase the policy networks to 3 layers with 1024 hidden units due to the high dimensional state space.
We use Relu activations for hidden layers and Adam optimizer. The batch size is 256, and the learning rate is $1e-4$. For discriminators, to satisfy the continuity assumption when deriving the corrective loss terms in Appendix~\ref{app:derivation_dmil}, the output is clipped to $[0.1,0.9]$ after sigmoid activation.
	
For both DMIL and D2MIL, we set $\alpha_\pi$ and $\alpha_f$ as $10$ across all tasks, which are found to achieve good performance. For D2MIL, $\eta=0.5$ is used in all experiments, and the additional weight hyperparameters $\beta_r$ and $\beta_o$ are set to $0.5$ in simulation experiments. In real-world experiments, due to large quality gap between the expert dataset and suboptimal human demonstrations, $\beta_o$ is set to $0.6$, and $\beta_r$ is set to $1-\beta_o=0.4$. Although DMIL and D2MIL contain several hyperparameters, we found them do not need careful tuning. Even using the same set of default parameters in different tasks, the model still provides good performance. \textcolor{black}{We summarize these hyperparameters in Table \ref{param} and provide evaluation and discussions on the different choices of hyperparameters in Appendix \ref{app:hyper_eval}.}

\subsection{Detailed Experiment Settings}\label{app:exp_details}
\textbf{D4RL Benchmark Experiments.}
In D4RL benchmark tasks under simulation environment, we use the medium and expert datasets in Mujoco and human dataset in Adroit of D4RL~\cite{d4rl} to conduct our experiments.
	There are 1 million samples in each expert or medium dataset for D4RL-MuJoCo tasks. We randomly sample 10\%, 5\% and 2\% of transitions from these MuJoCo datasets (correspond to 100,000, 50,000, 20,000 transitions) to evaluate policy performance under small datasets.  
	For Adroit tasks, there are only 5,000, 11,310 and 6,729 transitions in human datasets for Pen, Hammer and Door tasks respectively, which are already small compared with their high dimensionality in state space. Hence we directly use the original human datasets in our experiments.
	To evaluate the policy robustness, we randomly pick 20\% samples from previous constructed datasets and add a Gaussian noise with $\mathcal{N}(0,\sigma^2)$ on the states, where $\sigma$ stands for the standard deviation of each dimension of states in training dataset. As for the evaluation on D2MIL, we first sample 1\% trajectories (10,000 transitions) from D4RL-MuJoCo expert datasets, then sample $X$ proportion of these trajectories and combine them with 2\% medium dataset (20,000 transitions) to constitute the suboptimal dataset $\mathcal{D}_o$. The remaining 1-$X$ trajectories constitute the expert dataset $\mathcal{D}_e$. We term each task in different environments as exp-med-$X$. Detailed statistics of the datasets used in the experiments are summarized in Table~\ref{datasets}.


\begin{figure}[h]
\small
  \centering
  \subfloat[Hopper]{\includegraphics[width=0.15\linewidth]{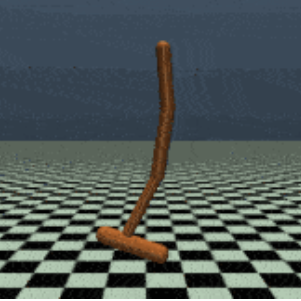}}
  \subfloat[Halfcheetah]{\includegraphics[width=0.15\linewidth]{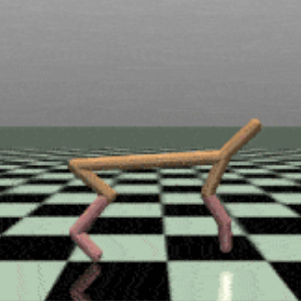}}
  \subfloat[Walker]{\includegraphics[width=0.15\linewidth]{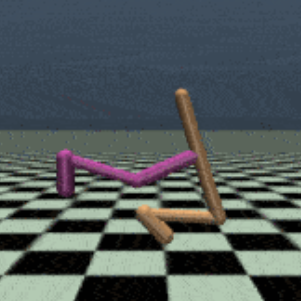}}
  \subfloat[Pen]{\includegraphics[width=0.15\linewidth]{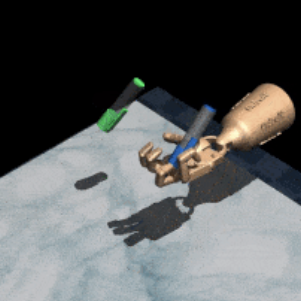}}
  \subfloat[Hammer]{\includegraphics[width=0.15\linewidth]{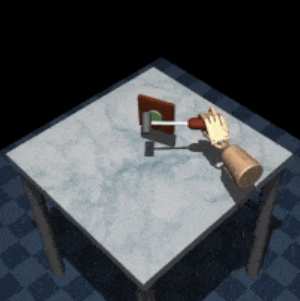}}
  \subfloat[Door]{\includegraphics[width=0.15\linewidth]{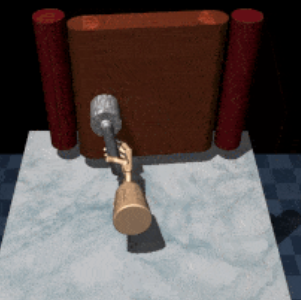}}
  \caption{Simulated tasks in D4RL benchmarks.} 
  \label{setting} 
\end{figure}

\textbf{Real-world Experiments.}
For real-world validation, we deploy our methods and baselines on a wheel-legged robot. The control action is the sum of the torque $\tau$ of the motors at the two wheels ($\frac{\tau}{2}$ for each). The control frequency of the robot is 200Hz. We elaborate the two task settings as follows:
	
	(1) \textbf{Standing still}: The state space of the robot is represented by $\mathbf{s} =(\theta, \dot \theta, x, \dot x)$,  where $\theta$ denotes the forward tilt angle of the body, $x$ is the displacement of the robot, $\dot \theta$ is angular velocity, and $\dot x$ is linear velocity. We collect datasets containing human controlled transitions of $(\mathbf{s}, a, \mathbf{s}', r, d)$, where $\mathbf{s}$ is the current state, $a$ is the torque of motors, $\mathbf{s}'$ is the next state, $r$ is the reward and $d$ is the flag of terminal. 
    During performance evaluation, we run all algorithms for 200,000 training steps and report the final results in the main text.
	
	(2) \textbf{Moving straight}: 
	The state space in this task is represented by $\mathbf{s} =(\theta, \dot \theta, \dot x)$, without the forementioned displacement $x$ since we only want to keep the velocity of the robot stable.
	Datasets contain human controlled transitions of $(\mathbf{s}, a, \mathbf{s}', r, d)$ when the robot moves forward. Our goal is to keep the robot at the target speed of 0.2m/s. 
During performance evaluation, we run all algorithm for 200,000 training steps and report the final results in the main text.

For each of the above two tasks, we collect 10,000 transition data from human demonstrations, which are about 50 second human control. As the actual control frequency of the robot is high (200Hz), human demonstrations can only be perceived as mediocre or suboptimal data. To evaluate the performance of D2MIL, we additionally collect very few transitions (140 transitions, less than 1 second's control) generated by a high quality Linear Quadratic Regulator (LQR) policy for the standing still task. We use such very small amount of expert data combined with human demonstrations to evaluate and compare the performance of D2MIL against baseline methods.

\section{Additional Experiment Results}\label{app:add_exps}
\subsection{Additional Comparative Evaluation Results}\label{app:add_exps_compare}
\textbf{Simulation Experiments on D4RL-MuJoCo Medium Datasets.}
We also evaluate DMIL on D4RL-MuJoCo medium-quality datasets, which are generated from a policy trained to approximately 1/3 the performance of an expert policy. The comparative results are shown in Table \ref{medium}. Due to the suboptimality in medium datasets, the gap between different methods is not as large as the experiments on expert data (Table~\ref{table1} in the main text). However, we can still observe that DMIL consistently outperforms other baselines in all tasks.

\begin{table}[t]
\small
\caption{Normalized scores for models trained on different proportion of D4RL MuJoCo-medium datasets. Results are averaged over 3 random seeds.}
\begin{adjustbox}{center}
\begin{tabular}{cccccccc}
\toprule
                             & Ratio & BC              & BC+d  &2-phase BC+d      & valueDICE &IQ-Learn  & DMIL            \\ \midrule
\multirow{3}{*}{Hopper-med}  & 10\%    & 46.26$\pm$8.69  & 47.55$\pm$7.56 &48.55$\pm$7.30  & \textbf{53.96$\pm$5.48}    &47.01$\pm$5.59                                & \textbf{53.72$\pm$8.78}  \\
                             & 5\%    & 43.31$\pm$8.81  & 45.19$\pm$7.86 &46.47$\pm$7.13  & 52.43$\pm$8.92    &43.88$\pm$5.67                              & \textbf{52.81$\pm$8.47}  \\
                             & 2\%    & 41.35$\pm$8.38  & 41.44$\pm$6.51 &46.07$\pm$6.87  & 51.43$\pm$6.48    &25.42$\pm$3.02                              & \textbf{52.89$\pm$8.42}  \\ \midrule
\multirow{3}{*}{Halfcheetah-med} & 10\%    & 41.58$\pm$1.69  & 41.12$\pm$1.49 &41.35$\pm$2.23  & 40.81$\pm$2.32   &40.36$\pm$1.92                                 & \textbf{41.86$\pm$2.19}  \\
                             & 5\%    & 40.46$\pm$2.61  & 40.47$\pm$1.65 &41.15$\pm$2.31  & 40.23$\pm$2.46    &36.66$\pm$4.27                                & \textbf{42.19$\pm$2.56}  \\
                             & 2\%    & 36.29$\pm$5.71  & 34.59$\pm$5.91 &39.37$\pm$3.46  & 37.21$\pm$1.89    &27.45$\pm$8.24                                & \textbf{41.26$\pm$1.61}  \\ \midrule
\multirow{3}{*}{Walker2d-med}  & 10\%    & 66.14$\pm$16.54 & 66.25$\pm$15.54 &68.08$\pm$15.28 & 47.11$\pm$3.55    &54.28$\pm$11.74                               & \textbf{71.66$\pm$12.51} \\
                             & 5\%    & 62.62$\pm$19.84 & 64.38$\pm$18.97 &64.95$\pm$18.13 & 37.86$\pm$8.99     &13.57$\pm$8.28                               & \textbf{67.51$\pm$15.75} \\
                             & 2\%    & 44.84$\pm$25.50 & 47.82$\pm$25.39 &59.52$\pm$21.00 & 33.35$\pm$6.11     &5.87$\pm$4.24                           & \textbf{62.25$\pm$17.05} \\ \bottomrule
\end{tabular}
\end{adjustbox}
\label{medium}
\end{table}

\textbf{Real-world Experiments for Scenarios with Additional Suboptimal Dataset.}
We also conduct real-world experiments on standing still task for D2MIL. In this setting, we collect 140 transitions generated from a high quality LQR expert policy. In particular, we consider two different sizes of expert dataset $\mathcal{D}_e$, one contains all the 140 transitions, the other contains only 1/10 of the data, 14 transitions. We also sample 5,000 transitions from the human demonstrations to constitute the suboptimal dataset $\mathcal{D}_o$. The amount of expert data, especially the second case, is extremely small compared with the suboptimal data, which requires the IL algorithm to maximally extract information from the suboptimal datset $\mathcal{D}_o$ for policy learning.

\begin{figure}[t]
		\small 
		\begin{minipage}[b]{1\linewidth}
			\centerline{\includegraphics[height=4.75cm]{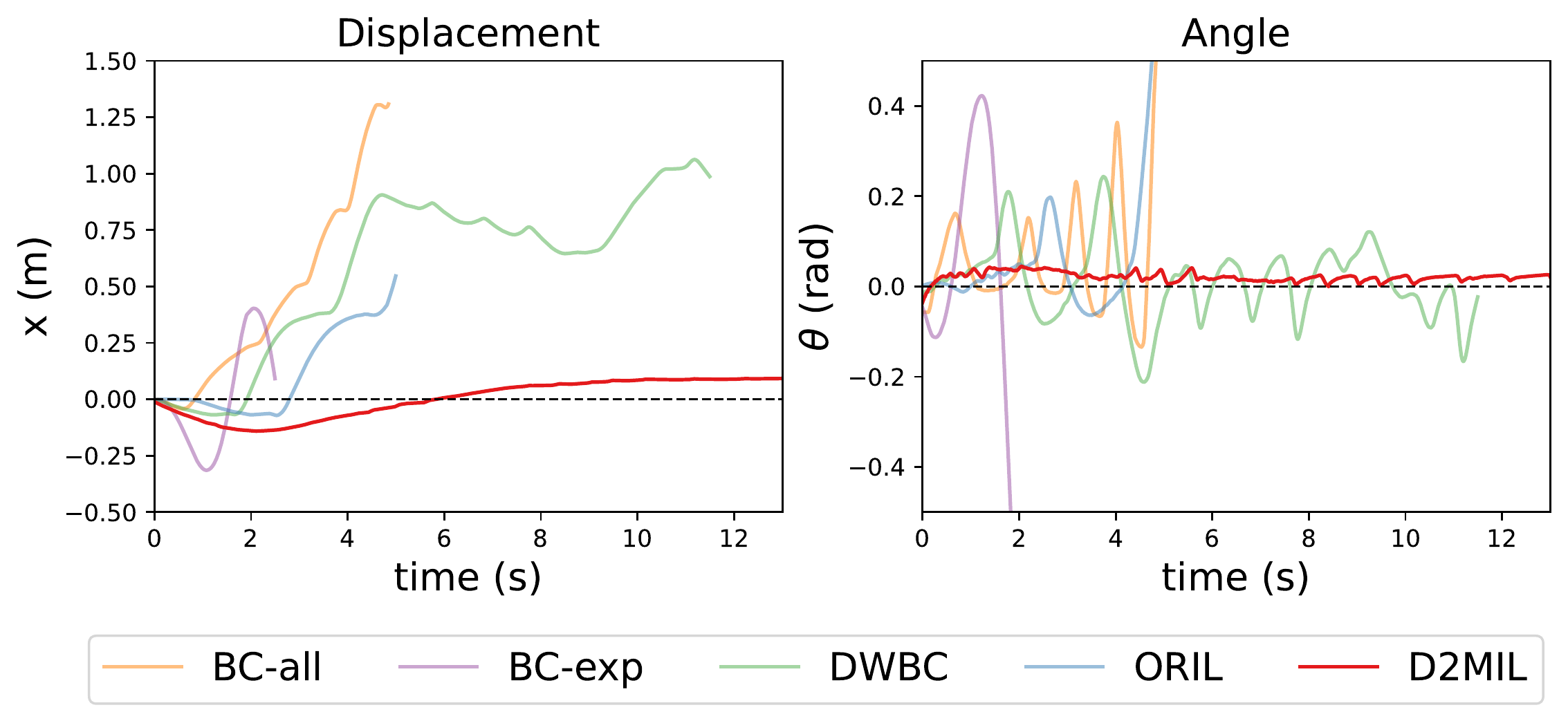}}
			\centerline{(a) D2MIL trained on 140 expert transitions and 5,000 suboptimal human demonstration transitions.}
		\end{minipage} 
		\vfill
		\begin{minipage}[b]{1\linewidth}
			\centerline{\includegraphics[height=4.75cm]{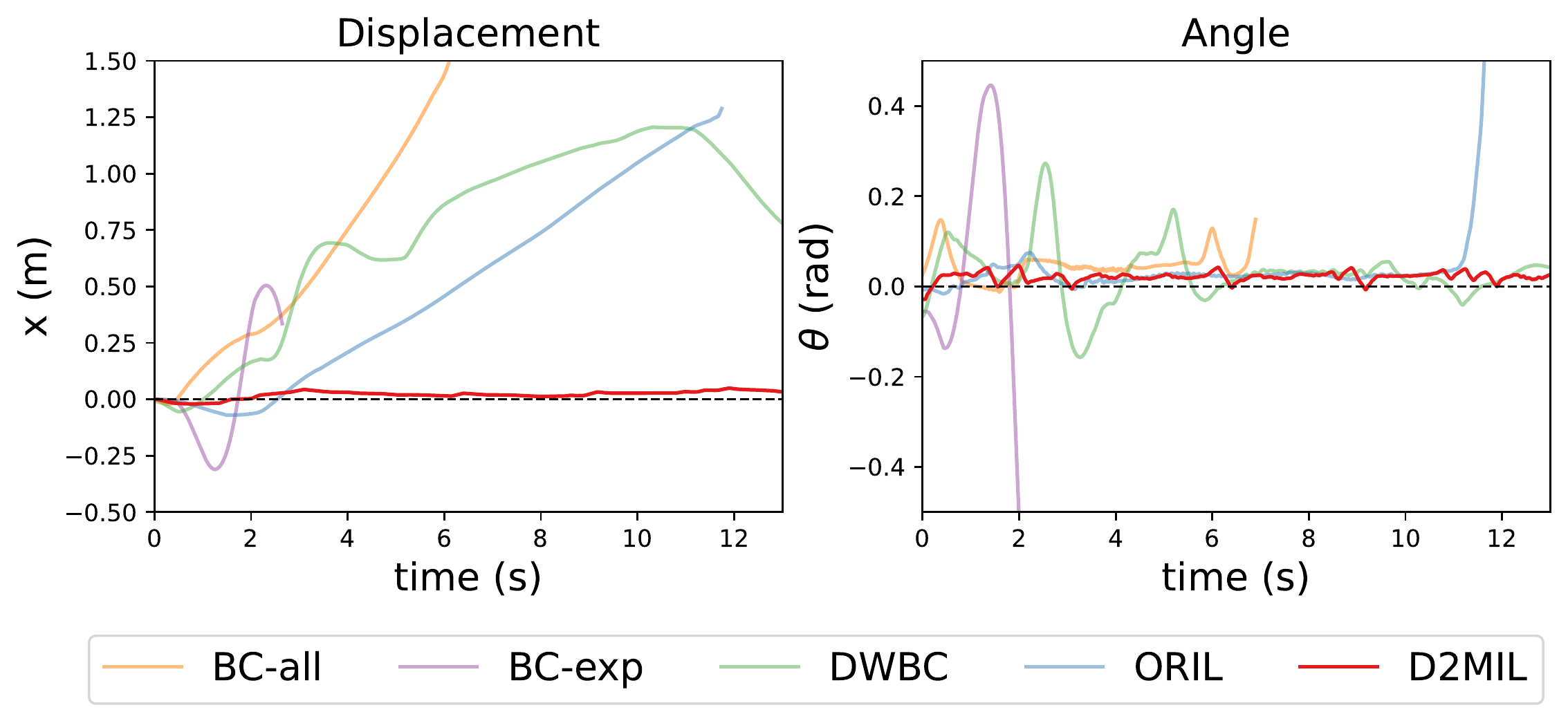}}
			\centerline{(b) D2MIL trained on 14 expert transitions and 5,000 suboptimal human demonstration transitions.}
		\end{minipage} 
\caption{\small Evaluation results of D2MIL on the standing still task on the real-world wheel-legged robot}
\label{robot-d2mil}
\end{figure}

The evaluation results are shown in Figure \ref{robot-d2mil}. Robot trained with BC-all, BC-exp and ORIL polices cannot maintain balance in both task settings. Although robot trained with DWBC can maintain a rather stable tilt angle, it fails to stay still and shows a slight drift. While for D2MIL, robot can stay in place and keep balance at the same time, indicating superior performance over other baselines.

	\subsection{Ablation on the Cooperative-yet-Adversarial Learning Scheme. }\label{app:add_exps_ablation}
	We conduct ablation study on D4RL-MuJoCO expert datasets to examine the benefits of introducing the proposed cooperative-yet-adversarial learning scheme in DMIL. This scheme has two ingredients, first is the incorporating element-wise loss information $\log\pi$ and $\log f$ into the discriminator $d$ to establish cooperative information sharing; the second is adding adversarial learning strategy between both $\pi$ and $f$ against $d$. To examine the impact of these ingredients, we evaluate the following baselines or variants of DMIL on MuJoCo expert and 20\% state noise datasets:
	\begin{itemize}[leftmargin=*,topsep=0pt]
		\item \textbf{DMIL-no-d-adv}: removing the coupling and the adversarial relationship between discriminator $d$ and dynamics model $f$. In this variant, we remove both the additional information $\log f$ from the inputs of $d$, as well as the corrective loss term $\mathcal{L}_f^{corr}$ of $f$ to remove its adversarial behavior on $d$.
		\item \textbf{DMIL-no-d-adv\&$\pi$-info}: on the basis of DMIL-no-d-adv, we further remove the additional information $\log \pi$ from the inputs of $d$. This removes the cooperative information sharing in DMIL, but we keep the corrective loss term $\mathcal{L}_{\pi}^{corr}$ of $\pi$ to enable discriminator-guided policy learning.
		\item \textbf{2-phase BC+d}: this baseline can be perceived as the reduction of DMIL that completely removes the cooperative-yet-adversarial learning scheme.
		\item \textbf{BC} and \textbf{BC+d}: minimal baselines without or with a dynamics model used for comparison.
	\end{itemize}
	
	The results are presented in Table \ref{ablation}. 
	From the results, we can see that without the cooperative-yet-adversarial learning scheme (BC, BC+d, 2-phase BC+d), the performance of imitation policy degenerates significantly on small datasets. When incorporating the adversarial relationship between policy $\pi$ and discriminator $d$ (DMIL, DMIL-no-d-adv\&$\pi$-info, DMIL-no-d-adv), the performance of policy is substantially improved under small dataset.
	As for DMIL-no-d-adv and DMIL-no-d-adv\&$\pi$-info that remove adversarial relationship between $f$ and $d$, they have similar performance with DMIL when the training data are sufficient, but suffer from noticeable performance drop when dataset is extremely small or contains noisy inputs. 
	On the contrary, DMIL can maintain nearly the same performance with reduced datasets as well as involvement of noisy data. Therefore, we can see that the cooperative-yet-adversarial learning scheme involving $\pi$, $f$ and $d$ indeed help with improving policy robustness and imitation performance.

\begin{table}[tb]
\small
\caption{Ablation study of DMIL on different proportion of D4RL-MuJoCo expert and 20\% state noise datasets.}
\begin{adjustbox}{center}
\begin{tabular}{cccccccc}
\toprule
                             & ratio & BC              & BC+d &2-phase BC+d       & DMIL-no-d-adv\&$\pi$-info &DMIL-no-d-adv  & DMIL            \\ \midrule
\multirow{3}{*}{Hopper}
                          & 10\%    & 83.52$\pm$30.58 & 100.59$\pm$13.21 &104.35$\pm$9.44 &110.58$\pm$1.26     & 110.14$\pm$1.92                           & \textbf{111.56$\pm$1.51} \\
                             & 5\%    & 73.35$\pm$37.04 & 94.82$\pm$19.72 &99.66$\pm$14.98  & 109.26$\pm$2.51      & 108.44$\pm$4.49                           & \textbf{111.14$\pm$1.83} \\
                             & 2\%    & 53.54$\pm$36.89 & 61.57$\pm$30.18 &88.24$\pm$25.63  & 105.45$\pm$10.46       & 103.99$\pm$11.26                           & \textbf{108.51$\pm$3.88} \\ \midrule
\multirow{3}{*}{Halfcheetah} & 10\%    & 90.64$\pm$2.21  & 89.71$\pm$2.88 &71.27$\pm$19.33  & 92.38$\pm$2.69   & 92.22$\pm$2.42                           & \textbf{92.69$\pm$1.82}  \\
                             & 5\%    & 82.90$\pm$11.71 & 76.40$\pm$16.94  &70.89$\pm$23.06  &88.19$\pm$7.77     & 88.26$\pm$6.46                           & \textbf{90.18$\pm$4.43}  \\
                             & 2\%    & 23.58$\pm$16.36 & 21.48$\pm$16.86  &57.48$\pm$25.63  &59.79$\pm$28.56     & 53.71$\pm$28.70                           & \textbf{76.87$\pm$15.31} \\ \midrule
\multirow{3}{*}{Walker2d}  & 10\%    & 105.36$\pm$4.38 & 107.61$\pm$1.14  &106.40$\pm$1.96  & 107.68$\pm$0.91   & \textbf{108.29$\pm$1.13}                           & 107.62$\pm$0.83 \\
                             & 5\%    & 103.21$\pm$7.81 & 105.42$\pm$3.93 &104.51$\pm$4.54  & 107.11$\pm$1.02      & 106.30$\pm$1.36                           & \textbf{107.89$\pm$0.71} \\
                             & 2\%    & 58.34$\pm$35.86 & 60.64$\pm$35.10 &86.71$\pm$21.20  &101.40$\pm$10.76       & 103.76$\pm$5.43                           & \textbf{105.55$\pm$4.42} \\ 
                             \midrule[1.5pt]
\multirow{3}{*}{Hopper+noise}
                          & 10\%    & 74.28$\pm$29.69 & 75.66$\pm$31.14 &100.32$\pm$15.21 &106.84$\pm$7.57 & 107.79$\pm$6.07  & \textbf{110.17$\pm$1.95}     \\
                             & 5\%    & 66.71$\pm$30.23 & 71.48$\pm$30.98 &93.21$\pm$22.28 &104.98$\pm$10.02 & 105.64$\pm$6.23  & \textbf{109.62$\pm$3.02}     \\
                             & 2\%    & 47.86$\pm$29.18 & 43.56$\pm$29.12 &59.63$\pm$33.40 &101.21$\pm$15.43 & 98.76$\pm$18.37  & \textbf{108.47$\pm$4.78} \\ \midrule
\multirow{3}{*}{Halfcheetah+noise} & 10\%    & 84.90$\pm$7.58  & 86.84$\pm$4.96 &71.56$\pm$23.06 & 88.17$\pm$7.51  & 88.13$\pm$7.93   & \textbf{88.42$\pm$6.88}  \\
                             & 5\%    & 68.63$\pm$20.45 & 66.87$\pm$21.61 &67.46$\pm$25.85 &74.76$\pm$18.82 & 73.38$\pm$20.94  & \textbf{74.56$\pm$19.24}  \\
                             & 2\%    & 58.21$\pm$24.17 & 23.79$\pm$22.31 &61.74$\pm$23.08 &64.83$\pm$27.91 & 65.58$\pm$26.11  & \textbf{73.14$\pm$18.01} \\ \midrule
\multirow{3}{*}{Walker2d+noise}  & 10\%    & 104.28$\pm$5.69 & 97.21$\pm$16.99 &102.84$\pm$8.37 &107.01$\pm$2.03 & 105.40$\pm$5.71  & \textbf{107.94$\pm$0.64} \\
                             & 5\%    & 89.84$\pm$20.52 & 91.86$\pm$23.72 &97.38$\pm$15.87 &103.39$\pm$7.85 & 100.61$\pm$12.79 & \textbf{105.89$\pm$3.92} \\
                             & 2\%    & 66.98$\pm$37.23 & 74.76$\pm$35.07 &92.01$\pm$22.61 &92.22$\pm$22.76 & 95.13$\pm$23.93  & \textbf{103.54$\pm$6.98} \\ 
                             \bottomrule
\end{tabular}
\end{adjustbox}
\label{ablation}
\end{table}

{\color{black}
\subsection{Discussion and Evaluations on Different Choices of Hyperparameters}
\label{app:hyper_eval}
In the proposed DMIL, the hyperparameters involved are $\alpha_\pi$ and $\alpha_f$, which are used to balance the impact of correction loss terms. In general cases, we can simply choose $\alpha_\pi=\alpha_f>1$. In all our experiments, the values of $\alpha_\pi$ and $\alpha_f$ are set to be 10 without tuning (see Table \ref{param}), as we find this choice already produces good model performance. To further verify their impact, we conducted additional experiments on Hopper tasks with 2\% expert data by setting $\alpha_\pi$ and $\alpha_f$ to different values, the results are presented below. It is found that these hyperparameters generally do not need careful tuning and produce similar performance.

\begin{table}[h]
\caption{Experimental results for different values of hyperparameters in DMIL}
\begin{adjustbox}{center}
\begin{tabular}{cccc}
\toprule
$\alpha_\pi$, $\alpha_f$ & 5               & 10              & 20              \\\midrule
Results                 & 106.07$\pm$7.86 & 108.51$\pm$3.88 & 105.79$\pm$8.61 \\
\bottomrule
\end{tabular}
\end{adjustbox}
\end{table}

For the extended D2MIL, although we have hyperparemeters $\alpha_\pi$, $\alpha_f$, $\eta$, $\beta_o$ and $\beta_r$ in the model, most of them do not need to be tuned. As in DMIL, we set $\alpha_\pi=\alpha_f=10$. We adopt $\eta=0.5$ as a constant, which is same as in ORIL~\cite{oril} and DWBC~\cite{dwbc}. In our implementation, we make $\beta_o+\beta_r=1$ to reduce the parameter numbers. $\beta_o$ and $\beta_r$ reflect the trade-off between the reliability and optimality of samples in the suboptimal dataset $\mathcal{D}_o$ and rollout data $\mathcal{D}_r$. The detailed discussion on the impact of $\beta_o$ and $\beta_r$ is presented in the last paragraph of Appendix~\ref{app:derivation_d2mil}. In practical scenarios, we suggest the practitioners just setting $\beta_r=\beta_o=0.5$, which generally leads to reasonably good performance. In our real-world experiments, due to the large quality gap between the expert dataset and suboptimal human demonstrations, we set $\beta_o$ to be slightly larger value ($\beta_o=0.6$, $\beta_r=1-\beta_o=0.4$).


Although DMIL and D2MIL contain several hyperparameters, the associated hyperparameter tuning effort during practical use is actually very minor. We use the same set of hyperparameters in most of our experiments without tuning. Moreover, we found that using the default hyperparameter values summarized in Table \ref{param} in most cases lead to good performance. This can be a particularly nice feature for DMIL and D2MIL in practical applications.

}

\subsection{Co-evolution of Models During the Learning Process }
\begin{figure}[t] 
	\centering
	\makebox[\textwidth][c]{\includegraphics[width=\textwidth]{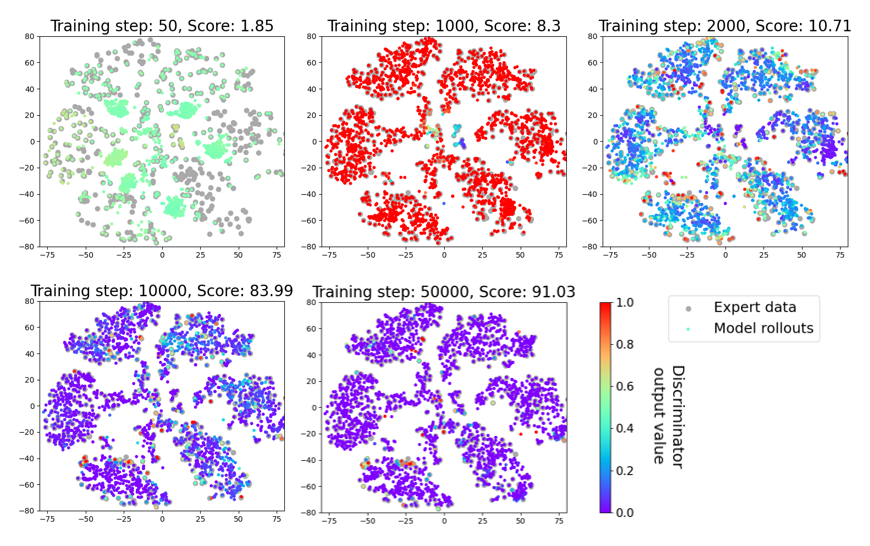}}
	\caption{\small TSNE visualization of the expert data and the generated rollout data under different stages of training on the Halfcheetah-5\% task. The color of rollouts points indicates the output value of the discriminator.
	}\label{process}
	\makebox[\textwidth][c]{\includegraphics[width=\textwidth]{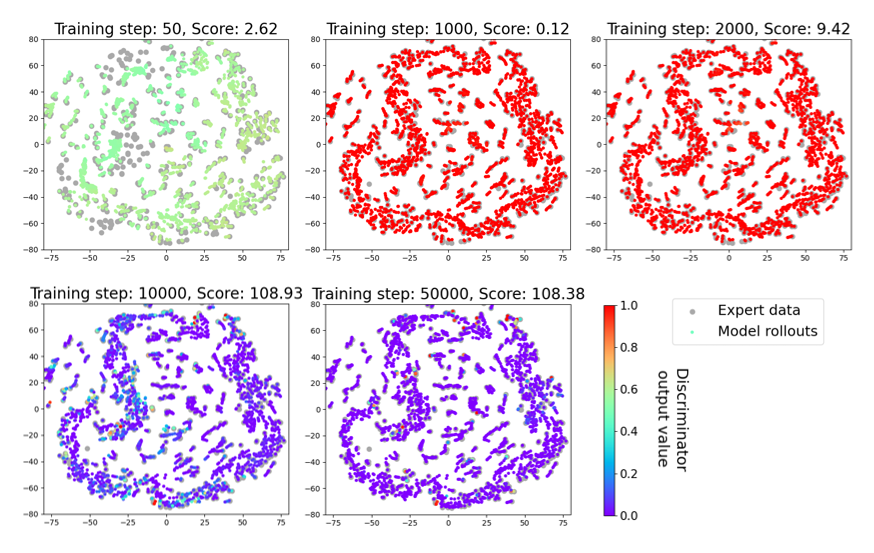}}
	\caption{\small TSNE visualization of the expert data and the generated rollout data under different stages of training on the Walker2d-5\% task. The color of rollouts points indicates the output value of the discriminator.
	}\label{process2}
\end{figure}

	To get a better understanding of our cooperative-yet-adversarial learning scheme in DMIL, we plot the TSNE visualization of generated model rollouts at different model training stages together with the original expert data in Figure \ref{process} and \ref{process2}. Moreover, we also plot the discriminator output values on these generated rollouts to examine how do the policy, dynamics model and discriminator co-evolve during training.
	We find that at the initial stage, the generated rollouts are inconsistent with expert data due to less well-learned policy, 
	and the discriminator $d$ is also incapable of discriminating the credibility of samples, which outputs around 0.5 for every rollout sample. As the training process continues and the policy is learned better, we can see that the generated rollouts start to align with the expert data, and the discriminator tends to believe most rollout data are reliable ($d\rightarrow 1$). However, at the later stage of training, as the discriminator is learned to be stronger, it can identify most of the generated rollouts are fake data ($d\rightarrow 0$). Under this stage, the policy will receive high learning weights only on few highly reliable samples, and the final imitation performance (illustrated as the average return scores in Figure \ref{process} and \ref{process2}) is gradually saturated.
	
	It is intriguing that above co-evolution pattern is almost universal across tasks, as observed in both Halfcheetah and Walker2d tasks with 5\% expert data. It is also worth noting that such a co-evolution pattern is very different from typical GAN-like methods. As in these approaches, the generator will eventually become stronger and the discriminator cannot tell whether the generated samples are real or fake (i.e., $d\rightarrow 0.5$). In DMIL, the discriminator $d$ can generally learned to be stronger compared with those in GAN-like method, due to additional cooperative information shared from $\pi$ and $f$ (i.e., adding $\log\pi$ and $\log f$ to the input of $d$). Moreover, since both $\pi$ and $f$ also optimize their own objectives in addition to enforcing the adversarial behavior on $d$, it is more likely the discriminator in DMIL can eventually distinguish most of the generated rollouts as fake. When such phenomenon occurs, it also suggests the saturation or convergence of the learning process.	

\subsection{Learning Curves}\label{app:add_exps_curvs}
\begin{figure}[h]
		\small 
		\begin{minipage}[b]{0.25\linewidth}
			\centerline{\includegraphics[height=4cm]{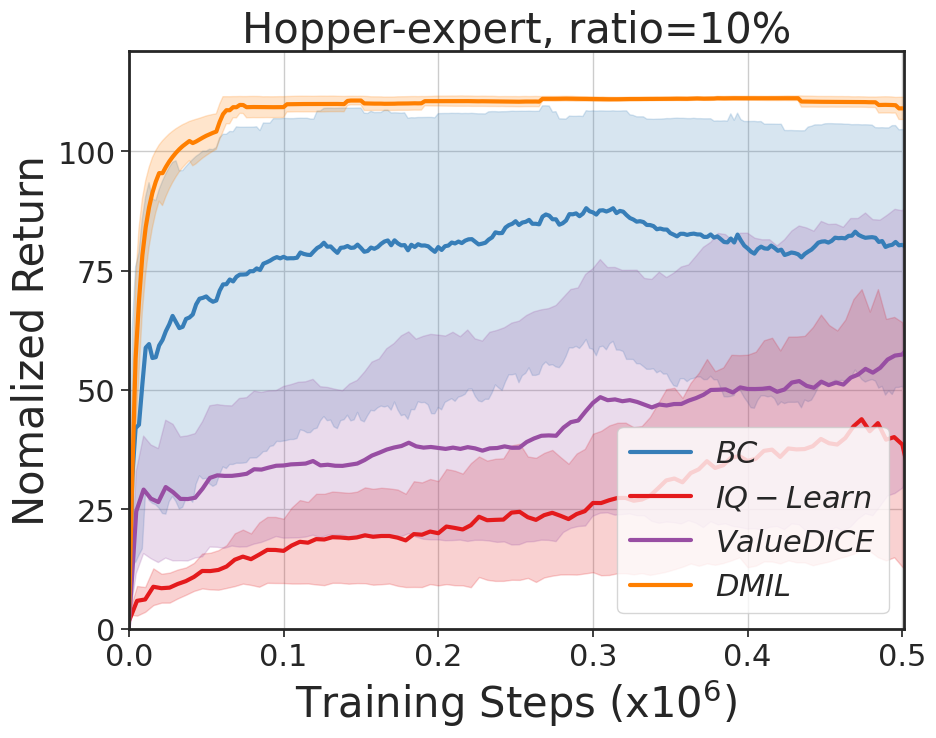}}
			\centerline{\includegraphics[height=4cm]{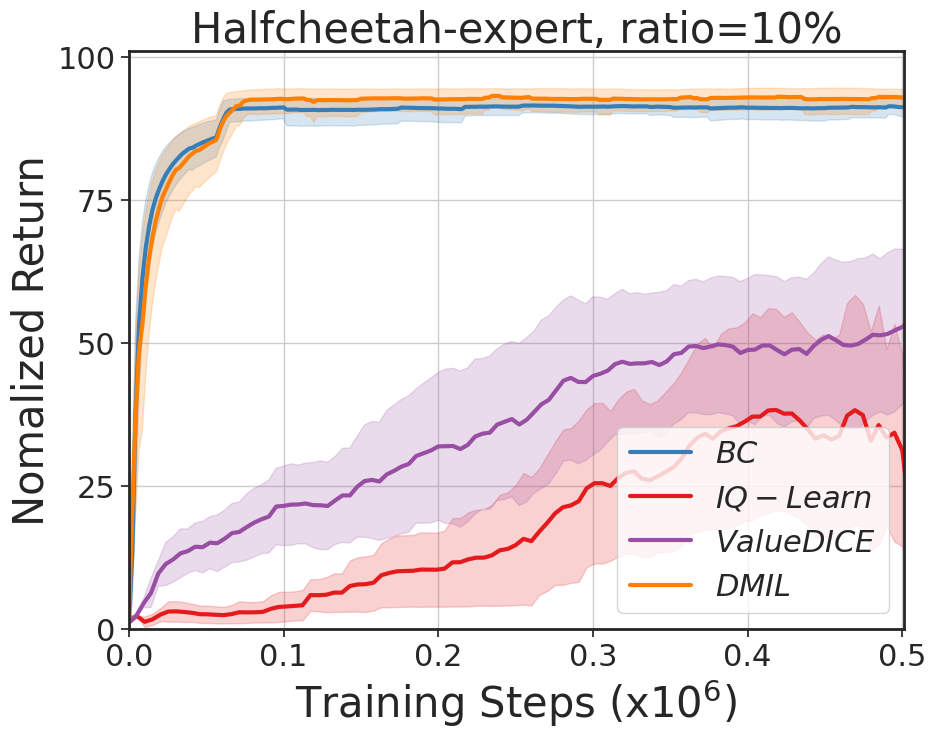}}
			\centerline{\includegraphics[height=4cm]{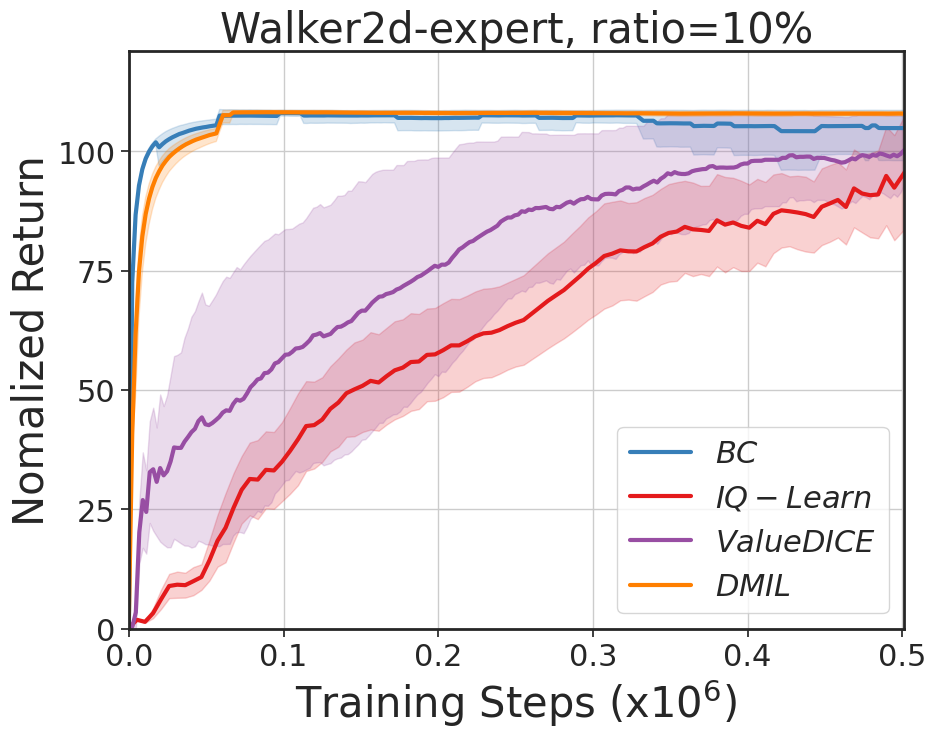}}
			\centerline{\includegraphics[height=4cm]{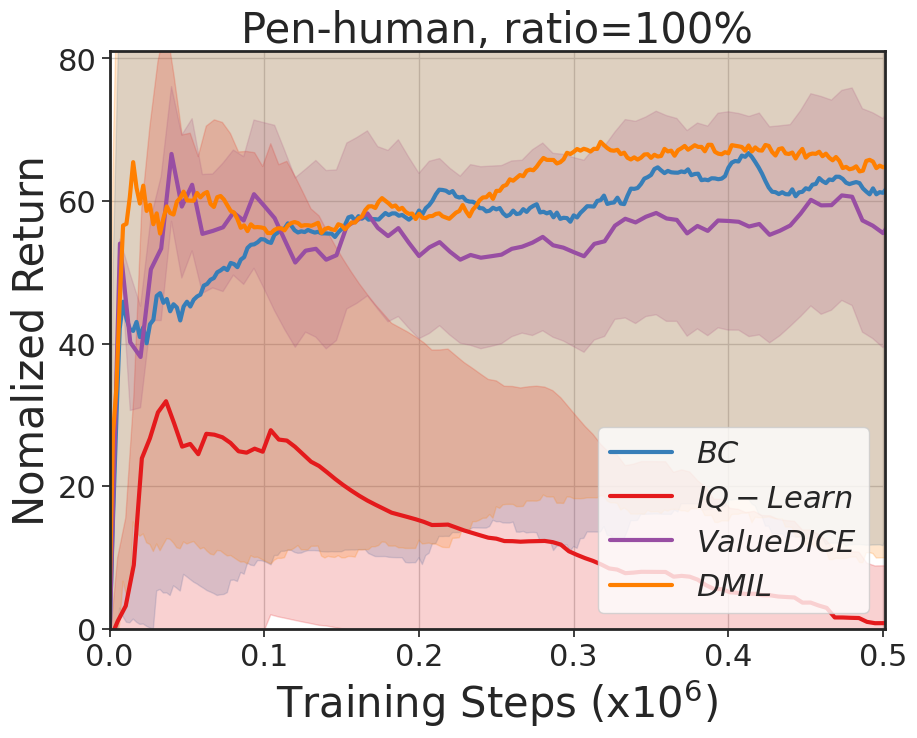}}
		\end{minipage} 
		\hfill
		\begin{minipage}[b]{0.25\linewidth}
			\centerline{\includegraphics[height=4cm]{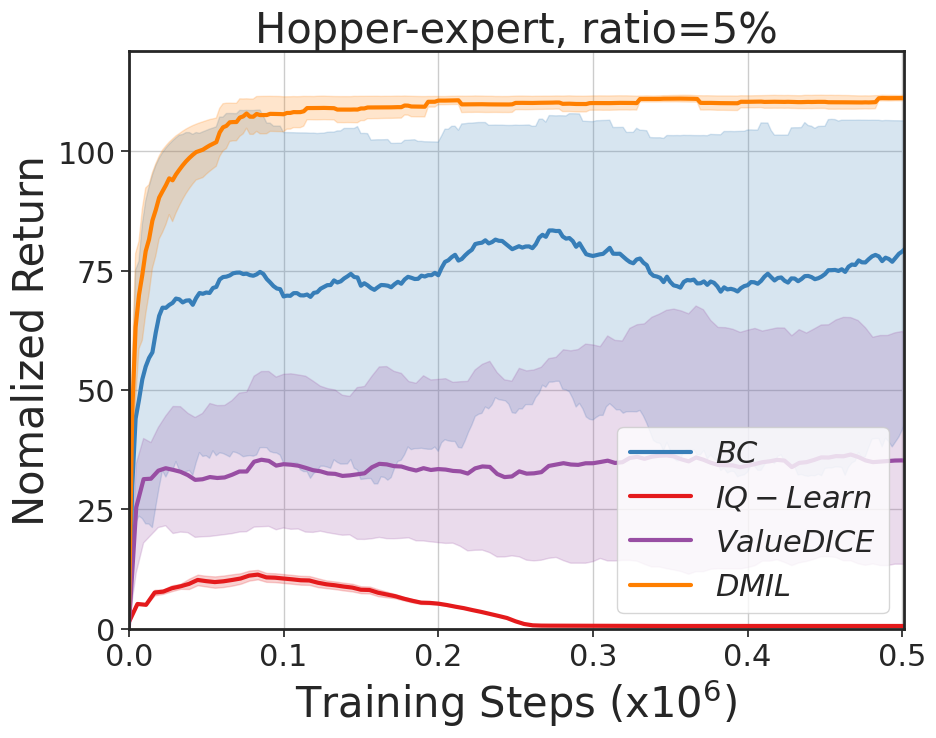}}
			\centerline{\includegraphics[height=4cm]{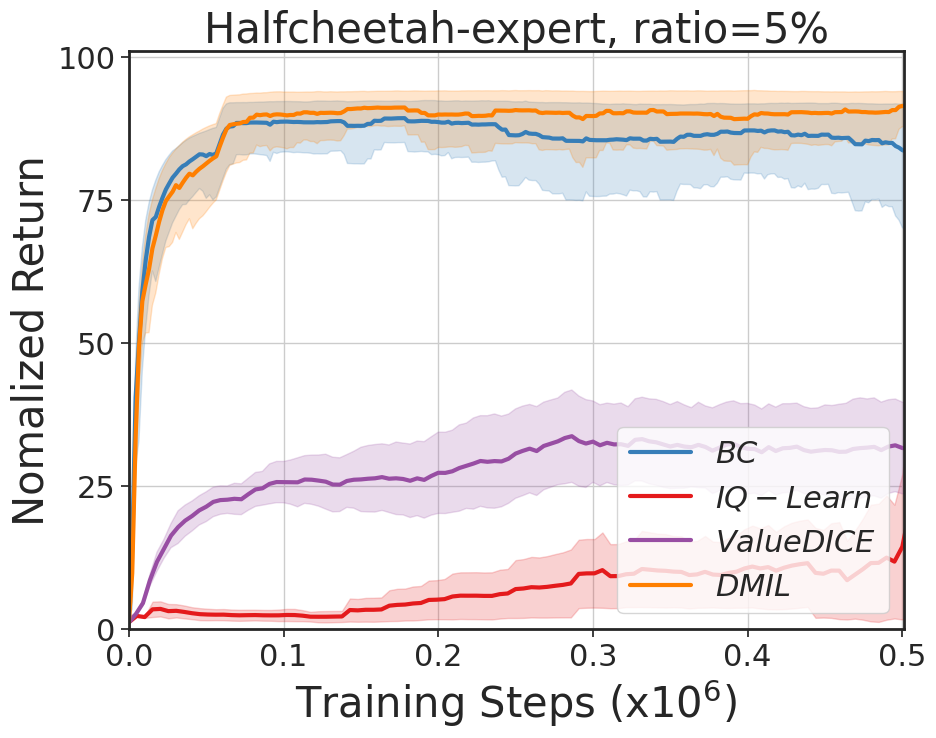}}
			\centerline{\includegraphics[height=4cm]{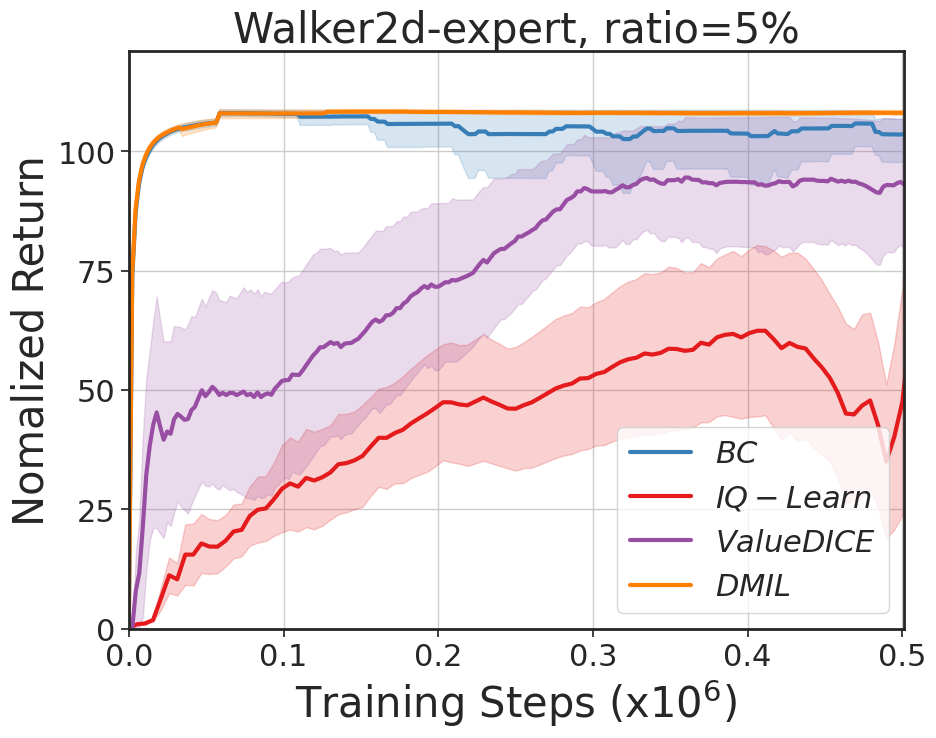}}
			\centerline{\includegraphics[height=4cm]{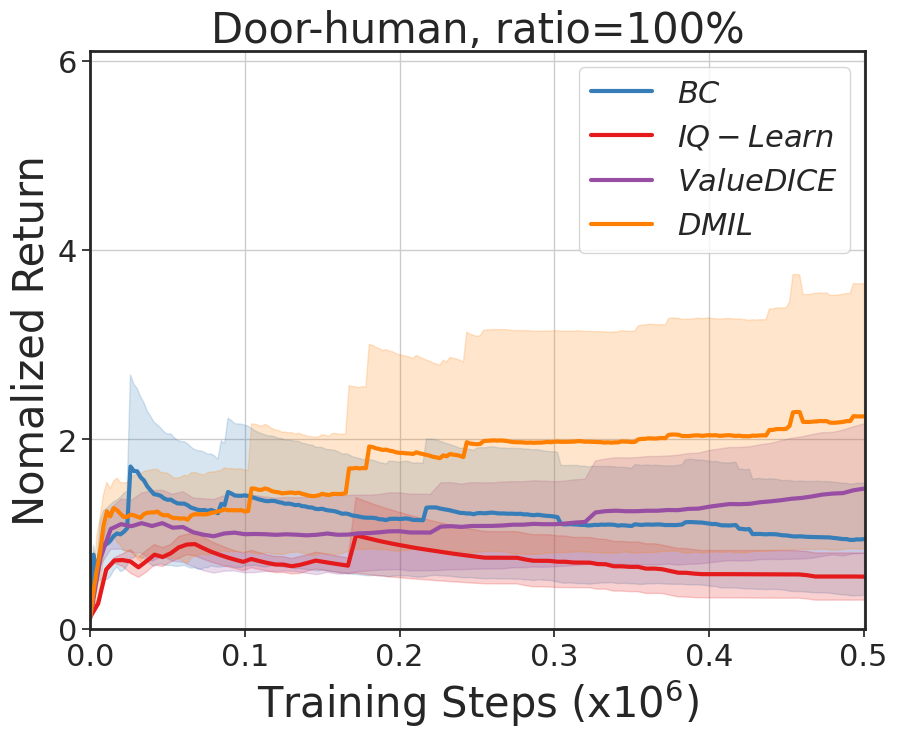}}
		\end{minipage} 
		\hfill
		\begin{minipage}[b]{0.25\linewidth}
			\centerline{\includegraphics[height=4cm]{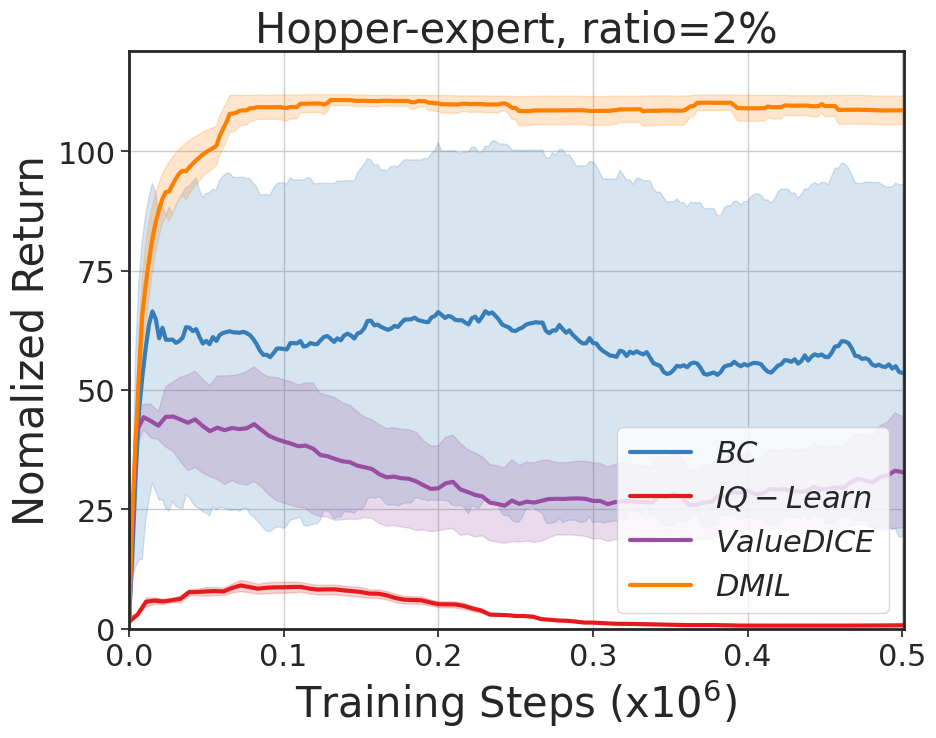}}
			\centerline{\includegraphics[height=4cm]{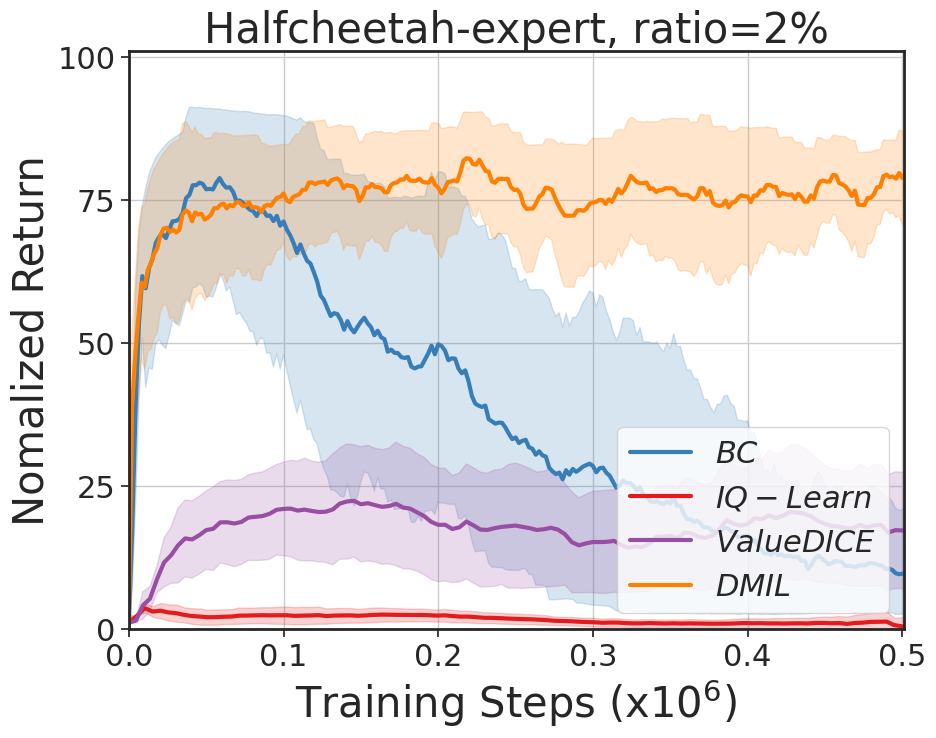}}
			\centerline{\includegraphics[height=4cm]{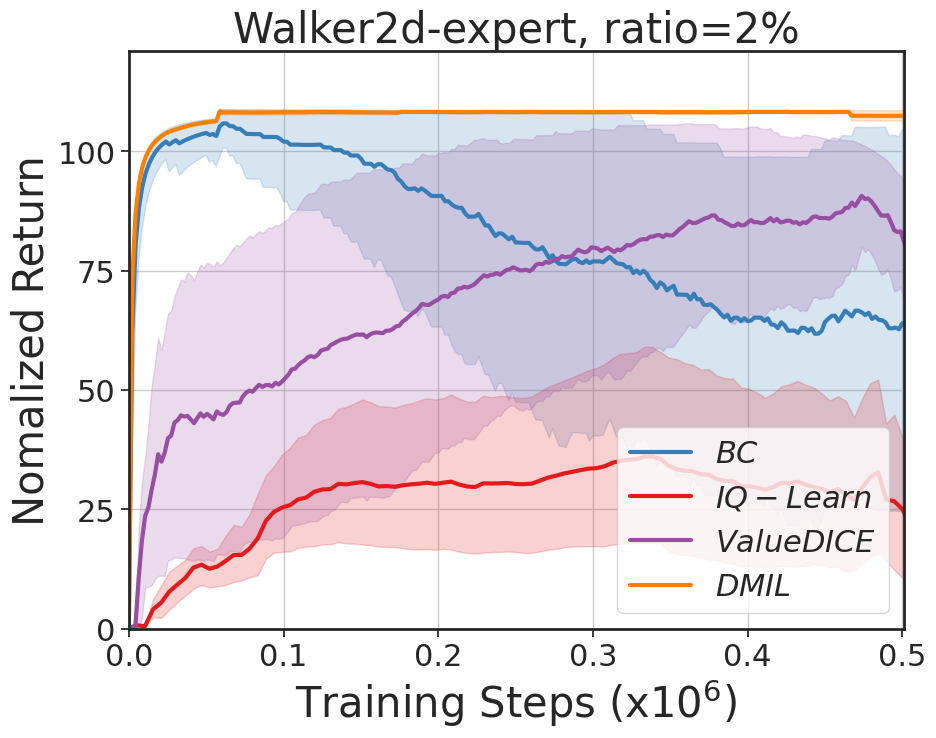}}
			\centerline{\includegraphics[height=4cm]{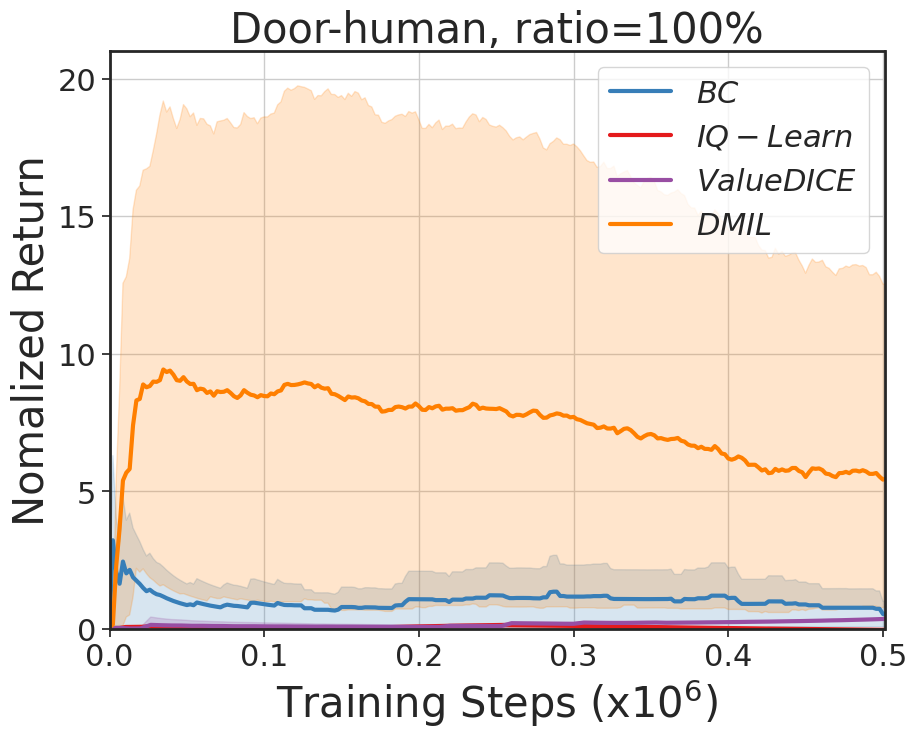}}
		\end{minipage} 
\caption{\small Learning curves of DMIL on D4RL benchmark tasks.}
\label{curve}
\end{figure}

The learning curves on D4RL benchmark tasks for DMIL are shown in Figure \ref{curve}. 

\end{document}